\pdfoutput=1

\documentclass[11pt]{article}

\usepackage[final]{acl}

\usepackage{times}
\usepackage{latexsym}

\usepackage[T1]{fontenc}

\usepackage[utf8]{inputenc}

\usepackage{microtype}

\usepackage{inconsolata}

\usepackage{graphicx}

\usepackage{mathtools}

\usepackage{booktabs}
\usepackage{multirow}
\usepackage{subcaption}
\usepackage{rotating}
\usepackage{makecell}
\usepackage{enumitem}

\usepackage{amsmath,amsfonts,bm}
\usepackage{xspace}

\def\eqref#1{equation~\ref{#1}}

\def\1{\bm{1}}

\DeclareMathAlphabet{\mathsfit}{\encodingdefault}{\sfdefault}{m}{sl}
\SetMathAlphabet{\mathsfit}{bold}{\encodingdefault}{\sfdefault}{bx}{n}

\newcommand{\ie}{\textit{i.e.}}
\newcommand{\eg}{\textit{e.g.}}

\newcommand{\langs}{\mathcal{L}}

\newcommand{\vocab}{\mathcal{V}}
\newcommand{\unembed}{W_U}

\newcommand{\repr}{\texttt{repr}}
\newcommand{\decargmax}{\texttt{dec-rmax}}
\newcommand{\dectopp}{\texttt{dec-topp}}

\newcommand{\startw}{\textsc{Start}(w)}
\newcommand{\startwlx}{\textsc{Start}_{\tau}(w_{\ell,x})}

\title{Lingua Franca or Probing Artifact?\\Rethinking Latent Language in Multilingual LLMs}

\author{
 \textbf{Deniz Bayazit},
 \textbf{Badr AlKhamissi},
 \textbf{Antoine Bosselut}
\\
 EPFL
\\
 \small{
   \textbf{Correspondence:} \texttt{\{deniz.bayazit,badr.alkhamissi,antoine.bosselut\}@epfl.ch}
 }
}

\begin{document}

\maketitle
\begin{abstract}
Latent language identification is often used to argue that multilingual language models route computation through language-specific states, such as English pivots. However, existing probes infer latent language from different signals, such as the geometry of hidden states or what can be decoded from intermediate representations.
Since such claims shape conclusions about how models share and route information across languages, we ask whether these probes measure the same phenomenon or expose distinct aspects of multilingual computation. 
We study this question across model families, training regimes, domains, tasks, checkpoints, and up to 27 languages. We find that identification probes systematically disagree: the GMM-based representation probe, which draws evidence from hidden state geometry, shows earlier cross-lingual mixing, whereas decoding-based probes, which rely on output-space decodability, retain sharper language-specific and more English-biased signals. These differences track model multilinguality and training progression, but are comparatively stable across domains. 
Our results suggest a more cautious interpretation of latent language identification, where current probes expose different aspects of multilingual processing, rather than directly revealing a single internal \textit{lingua franca}.\footnote{The code is available at: \\
{\url{https://github.com/bayazitdeniz/latent-lid}}}
\end{abstract}
\section{Introduction}

\begin{figure}[t]
\centering
\includegraphics[width=1.0\linewidth]{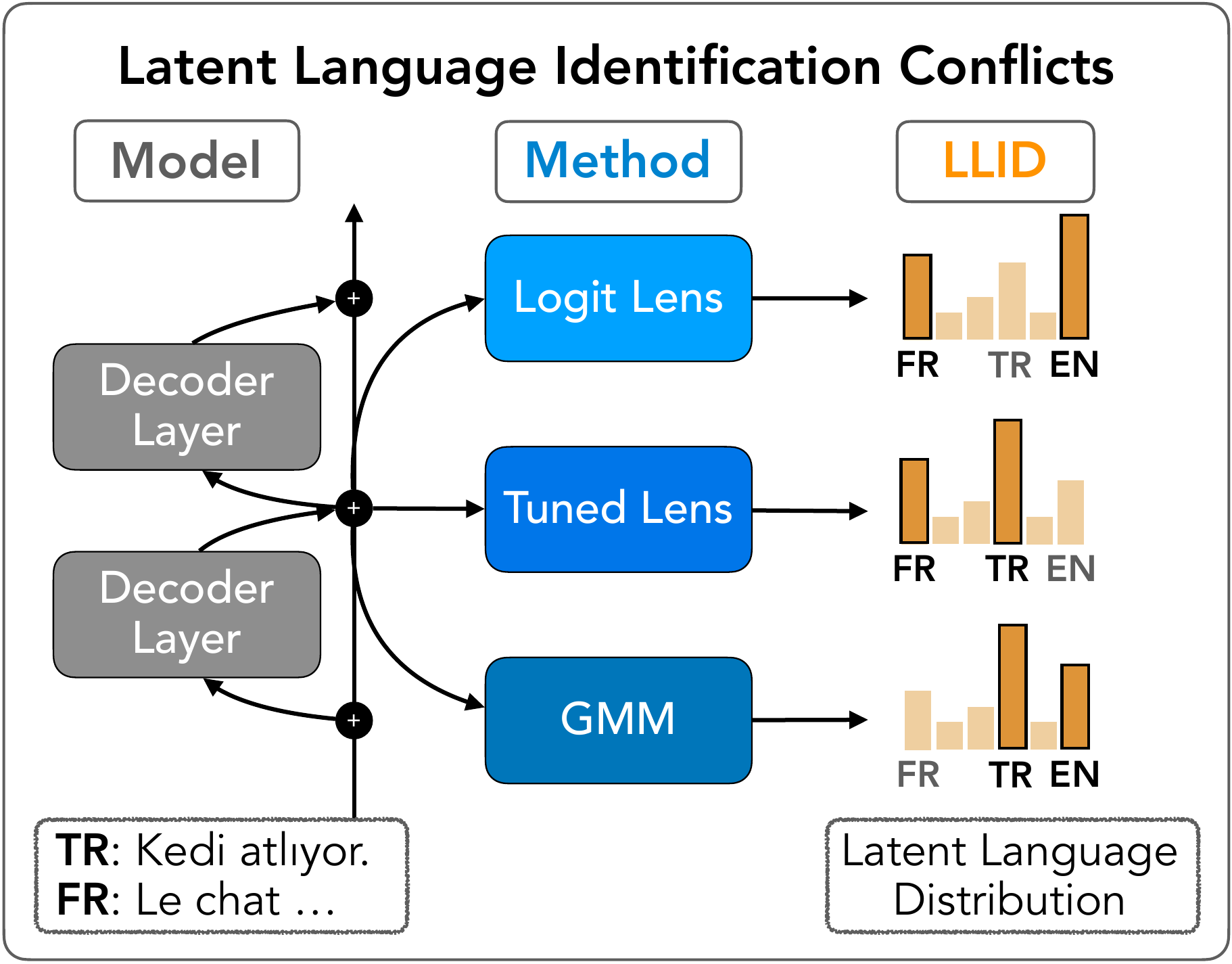}
\caption{\textbf{Do multilingual LLMs work in English? It depends on how you ask.} We probe the same intermediate hidden states of a multilingual LLM with three probes: logitlens, tuned lens, and a representation-based GMM, and recover three different latent language distributions over the task's languages (\textsc{tr}, \textsc{fr}), and other candidate languages like English (\textsc{en}). We study when and why these probes disagree across models, training regimes, and tasks.}
\label{fig:llid-overview}
\vspace{-1em}
\end{figure}

Multilingual language models are often assumed to share information across languages by mapping them into common or partially language-neutral representations \citep{foroutan2022discovering, dumas2025separatingtonguefromthought}. Yet the form of this sharing remains poorly understood. One influential hypothesis is that multilingual models rely on a \emph{latent language}: an intermediate language-like regime, sometimes described as an internal pivot or \textit{lingua franca}, through which the model routes meaning before producing an output. \citet{wendler2024llamas} provide evidence for such behavior in English-dominated models like LLaMA-2 \citep{touvron2023llama2}, but newer models trained on more balanced multilingual data may behave differently \citep{ustun2024aya,foroutan2025multilingdatamixtures,schut2025mutlilingual}. This phenomenon raises a broader question: \textit{what part of a latent language estimate reflects the model's internal computation, and what part reflects the assumptions of the diagnostic method}?

\begin{figure}
    \centering
    \includegraphics[width=1\linewidth]{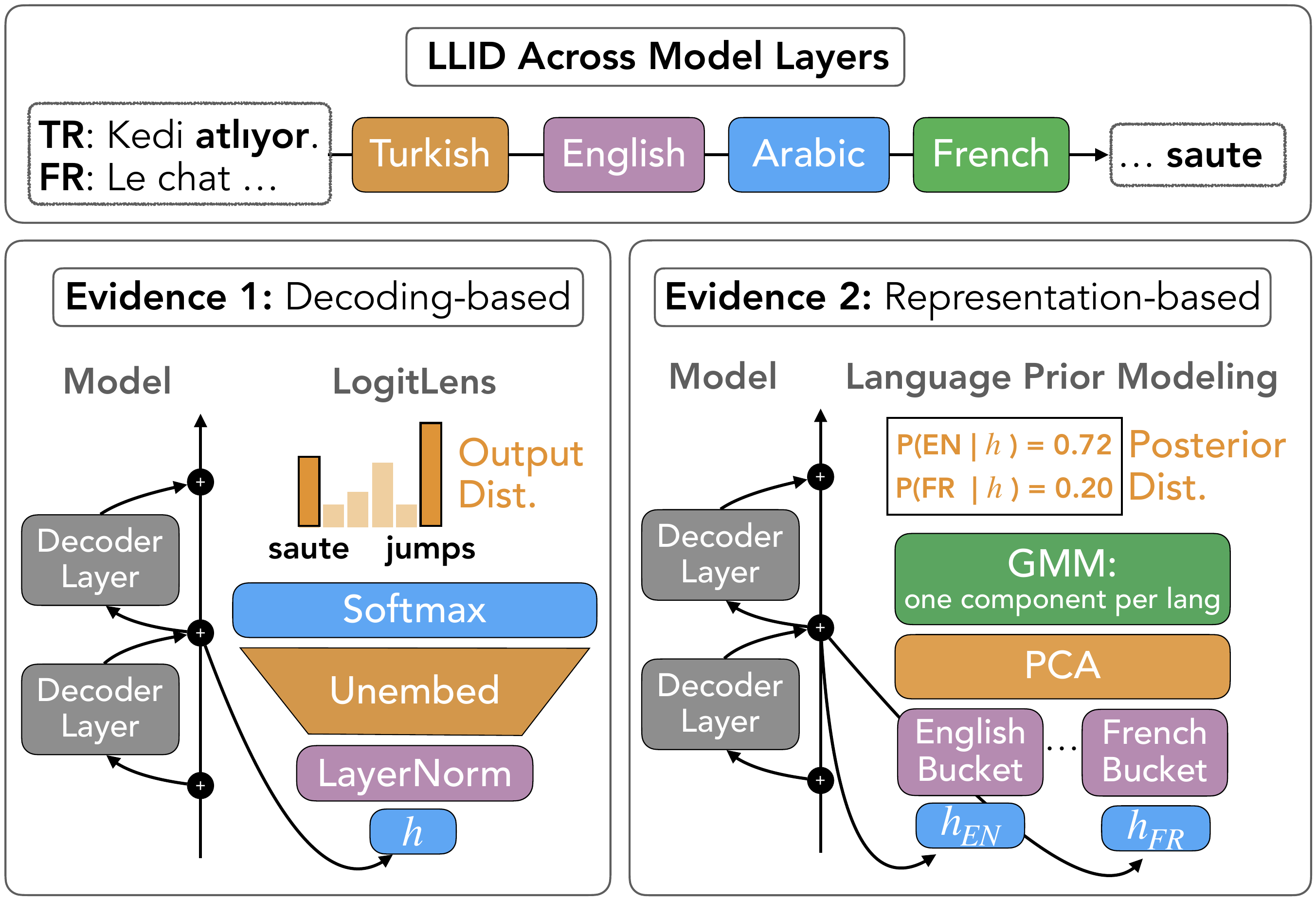}
    \caption{\textbf{Two families of latent language identification (LLID) probes.} Given an intermediate hidden state $h$, \textit{decoding-based} methods (left) project $h$ through the model's unembedding (optionally via a tuned lens) and derive language scores from the resulting vocabulary distribution. \textit{Representation-based} methods (right) instead compare $h$ with language-conditioned activation distributions, here a per-layer GMM with one component per candidate language. The two families operate on the same $h$ but extract different evidence: output-space decodability versus activation-space geometry.}
    \label{fig:probe-families}
    \vspace{-1em}
\end{figure}

Interpreting such probes is challenging because latent language identification (LLID) is inferred rather than directly observed. As illustrated in Fig.~\ref{fig:llid-overview}, a latent language estimate depends on what kind of evidence is extracted from the model. Prior work relies on two broad families (Fig.~\ref{fig:probe-families}): \textit{representation-based} probes, which infer language from the geometry of hidden states \citep{shani2025language}, and \textit{decoding-based} probes, which infer language by projecting intermediate hidden states through the model’s output head \citep{wendler2024llamas,zhong2025language}. These families make different assumptions about what it means for a model to internally use a language, and it is unclear whether they identify a common mechanism or expose different processes in multilingual encoding and decoding.

In this work, we treat LLID as a measurement problem. Instead of asking whether multilingual LLMs ``think in English,'' we ask whether latent language estimates reflect model properties, such as multilinguality and training progression, or diagnostic choices, such as representation geometry versus output-space decodability. 
To answer this, we compare a GMM-based representation probe and decoding probes across controlled and open-ended multilingual tasks. We first test the internal consistency and generalization of decoding-based LLID across controlled and open-ended settings (\S\ref{sec:synthetic-reproduced}–\ref{sec:open-ended}), before asking whether representation-based LLID recovers the same patterns under matched conditions (\S\ref{sec:repr-vs-dec}). We then trace how these estimates change across domains, training regimes, and checkpoints.

Our analysis shows that LLID probes should not be treated as interchangeable measurements of a single phenomenon. The GMM representation probe reveals earlier cross-lingual mixing and weaker English dominance, while decoding-based probes retain sharper language-specific and more English-biased signals. This pattern recurs across our analyses, but the layerwise behavior of each probe is model-dependent. For a given probe, where these effects emerge in the layer stack and how they change across layers vary with model multilinguality and over the course of model pretraining, although multilinguality alone does not determine them. By contrast, domain and language-inventory changes shift LLID estimates only modestly. Rather than supporting strong claims about an internal \textit{lingua franca}, each diagnostic provides a partial view of multilingual computation, shaped by the evidence it extracts from hidden states or decoded outputs.

\section{A Framework for Comparing Latent Language Probes}
\label{sec:framework}

\subsection{Task Definition: Latent Language Identification}

Let $\langs$ be a finite set of candidate languages, and let $h \in \mathbb{R}^d$ denote a hidden state extracted from a given model layer and sequence positions. We represent latent language with a variable $Z \in \langs$ and estimate its conditional distribution given $h$ as:
\begin{equation*}
q(\ell) \approx \Pr(Z=\ell \mid h), \quad \ell \in \langs
\end{equation*}
As the variable $Z$ is not directly observed, we treat $q$ as a diagnostic estimate, not as ground-truth evidence of an internal language variable. Different LLID estimators may therefore produce different distributions for the same prompt and layer. In practice, each estimator constructs $q$ by selecting one or more token positions and aggregating the resulting language evidence (\S\ref{sec:setup}).

\subsection{Two Kinds of Evidence for LLID and Their Limits}
\label{sec:framework-evidence}
Existing LLID methods differ mainly in the evidence they use to construct $q$. We distinguish two families. Representation-based methods infer language from where a hidden state lies in activation space relative to language-conditioned structure learned from multilingual data. Decoding-based methods infer language from what can be read out from that hidden state through the model's output vocabulary.

\paragraph{Representation-based LLID.}
This approach compares a hidden state $h$ to language-conditioned structure learned from multilingual data, such as GMM posteriors \citep{shani2025language}, language subspaces \citep{zhao2025languagereasoningdisentanglement}, or neuron-level indicators \citep{zeng2025linguafranca}. This comparison yields per-language scores, which are then aggregated into a layerwise distribution $q$. 

In our experiments, we fit a per-layer GMM with one language-conditioned component per candidate language using hidden states from examples in each language, and use the resulting component posteriors as the layerwise language distribution $q$.

A common concern for this approach is that learned structure may conflate language identity with correlated factors such as script, tokenization, or domain. In our experiments, however, we find consistent behavior across QA domains such as STEM and Arts \& Humanities (\S\ref{sec:domain-specific}).

\paragraph{Decoding-based LLID.}
These approaches estimate latent language by asking what language can be decoded from an intermediate layer. Let $\vocab$ be the model's vocabulary and let $\unembed \in \mathbb{R}^{|\vocab| \times d}$ be the model's unembedding matrix. Given any hidden state $h \in \mathbb{R}^d$, after final layer normalization, we form an intermediate vocabulary distribution:
{\setlength{\abovedisplayskip}{6pt}
 \setlength{\belowdisplayskip}{6pt}
\begin{equation*}
    p := \mathrm{softmax}(\unembed h)
\end{equation*}}
Applying the model’s unmodified output head directly to an intermediate hidden state is known as the logitlens \citep{nostalgebraist2020logitlens}, and we refer to this direct projection as the raw logitlens. Because intermediate representations may not align with those expected by the final output head, we also consider the tuned lens \citep{belrose2025tunedlens}, which learns a per-layer affine transformation to predict the final-layer hidden state from an intermediate one before applying the output head.

Prior work maps $p$ to language scores $q$ in several ways, such as summing probability over valid first-token prefixes of language-specific target words \citep{wendler2024llamas,dumas2025separatingtonguefromthought}, scoring the probability of multi-token language-specific answer sequences \citep{zhong2025language}, or decoding one or more tokens from $p$ and classifying the result with an external language identification (LID) model
\citep{ozaki2025llmsthinkonelanguage}. 
In our experiments, for tasks where the completion is known, we sum the probability over valid first-tokens, whereas if the task is open-ended, we use two rollout-based variants (argmax and top-$p$) where short continuations decoded from intermediate layers are passed to a LID classifier \citep{kargaran2023glotlid}.

Decoding-based LLID methods assume that the final unembedding remains meaningful at intermediate layers, which may not hold in earlier layers. A further concern is that token-level heuristics such as target-string matching require non-trivial design choices (\eg{}, gathering valid completions) to yield a meaningful distribution over $\langs$.

\subsection{Metrics to Compare LLID Estimates}
LLID estimates vary along several dimensions that cannot be captured by a single metric: distributions can be confident or diffuse, while their dominant language can match the task-relevant language or pivot to another one. Furthermore, two estimators can agree or disagree. We therefore use a small set of complementary metrics. Unless otherwise stated, each is computed per estimator and averaged over prompts at each layer.

\paragraph{Entropy.}
Entropy, $H=-\sum_{\ell\in\langs} q(\ell)\log q(\ell)$, measures how diffuse the LLID distribution is; lower values indicate a peaky estimate, while higher values indicate mixing across languages.

\paragraph{Dominance.} We measure dominance with $D=\max_{\ell\in\langs} q(\ell)$, which is high when most probability mass is assigned to a single language.

\paragraph{Pivot rate.}
Pivot rate is the fraction of prompts whose dominant estimated language differs from the task-relevant language $\ell_x$ (\eg{}, the source or target language in translation):
{\setlength{\abovedisplayskip}{6pt}
 \setlength{\belowdisplayskip}{6pt}
\begin{equation*}
P_x = \mathbb{I}\big[
    \arg\max_{\ell \in \langs} q(\ell) \neq \ell_x
    \big]
\end{equation*}}

\paragraph{Agreement.}
Given two LLID estimators ${a}$ and ${b}$, agreement measures how often the estimators assign the same dominant language:
{\setlength{\abovedisplayskip}{6pt}
 \setlength{\belowdisplayskip}{6pt}
\begin{equation*}
A_x
=
\mathbb{I}\big[
\arg\max_{\ell \in \langs} q_{a}(\ell)
=
\arg\max_{\ell \in \langs} q_{b}(\ell)
\big]
\end{equation*}}
\par\noindent In our experiments, this is used to compare decoding-based and representation-based LLIDs. 
\section{Experimental Setup}
\label{sec:setup}

\subsection{Models \& Measuring Multilinguality}
We evaluate up to 12 autoregressive language models spanning English-centric baselines, multilingual base models, and instruction-tuned variants. Because training mixtures are not fully available, we consider the following multilinguality proxies: zero-shot multilingual QA accuracy, cross-lingual perplexity, and byte-normalized likelihood. Unless otherwise stated, cross-model plots order models by increasing zero-shot macro-accuracy on \mbox{INCLUDE} \citep{romanou2024include}. Model details, scores, and rankings are in Appendix~\ref{sec:appendix_model}.

\subsection{Evaluation Regimes}
\label{sec:eval-regimes}

We evaluate LLID in two settings: (1) a controlled regime where ground-truth completions are available, and (2) an open-ended regime that tests whether the same patterns hold in natural text. The exact language inventory, \ie{}, the set of candidate languages included in each setting, is listed in Table~\ref{tab:dataset-language-inventories} in Appendix~\ref{sec:appendix_dataset}.

\paragraph{Controlled Target-String Evaluation.}
We use three synthetic task formats: copy, cloze, and translation. Copy asks the model to reproduce a concept word in the same language, cloze asks the model to complete a masked context, and translation asks for the concept word in a specified target language. These examples build on prior target-word settings for latent language analysis \citep{wendler2024llamas,dumas2025separatingtonguefromthought,zhong2025language}. For each prompt $x$ and candidate language $\ell$, let $w_{\ell,x}$ be the language-$\ell$ answer. Following \citeauthor{wendler2024llamas}, let 
$\startwlx \subseteq V$ 
denote the set of vocabulary tokens under tokenizer $\tau$ that can begin $w_{\ell,x}$. For simplicity, we fix $x$, $\ell$, and $\tau$ and write $\startw$. The controlled decoding-based language score is then defined as:
\[
s_x(\ell)
=
\sum_{\mathclap{t \in \startw}}
P_\theta(x_{n+1}=t \mid h_n),
\]
\par\noindent where $h_n$ is the hidden state at the final prompt token. We refer to this summed mass as the $\startw$ probability. When a language distribution is needed, we normalize $s_x$ over the candidate set $\langs$ to obtain $q_x$. Construction details for $\startw$ and candidate language completions are in Appendix~\ref{sec:appendix_target_string_construction}.

\paragraph{Open-Ended Evaluation.}

For natural-language prompts without constrained completions, we summarize both representation-based and decoding-based LLID at the midpoint of the prompt and aggregate over a forward window of five positions. We evaluate on PUD sentence data \citep{zeman-etal-2017-conll} and INCLUDE question-answering prompts \citep{romanou2024include}, using PUD variants to vary the language inventory and INCLUDE domains to test domain sensitivity. Exact dataset composition and additional evaluation details are in Appendix~\ref{sec:appendix_openended_eval}.

\begin{figure*}[th]
    \centering
    \includegraphics[width=\textwidth]{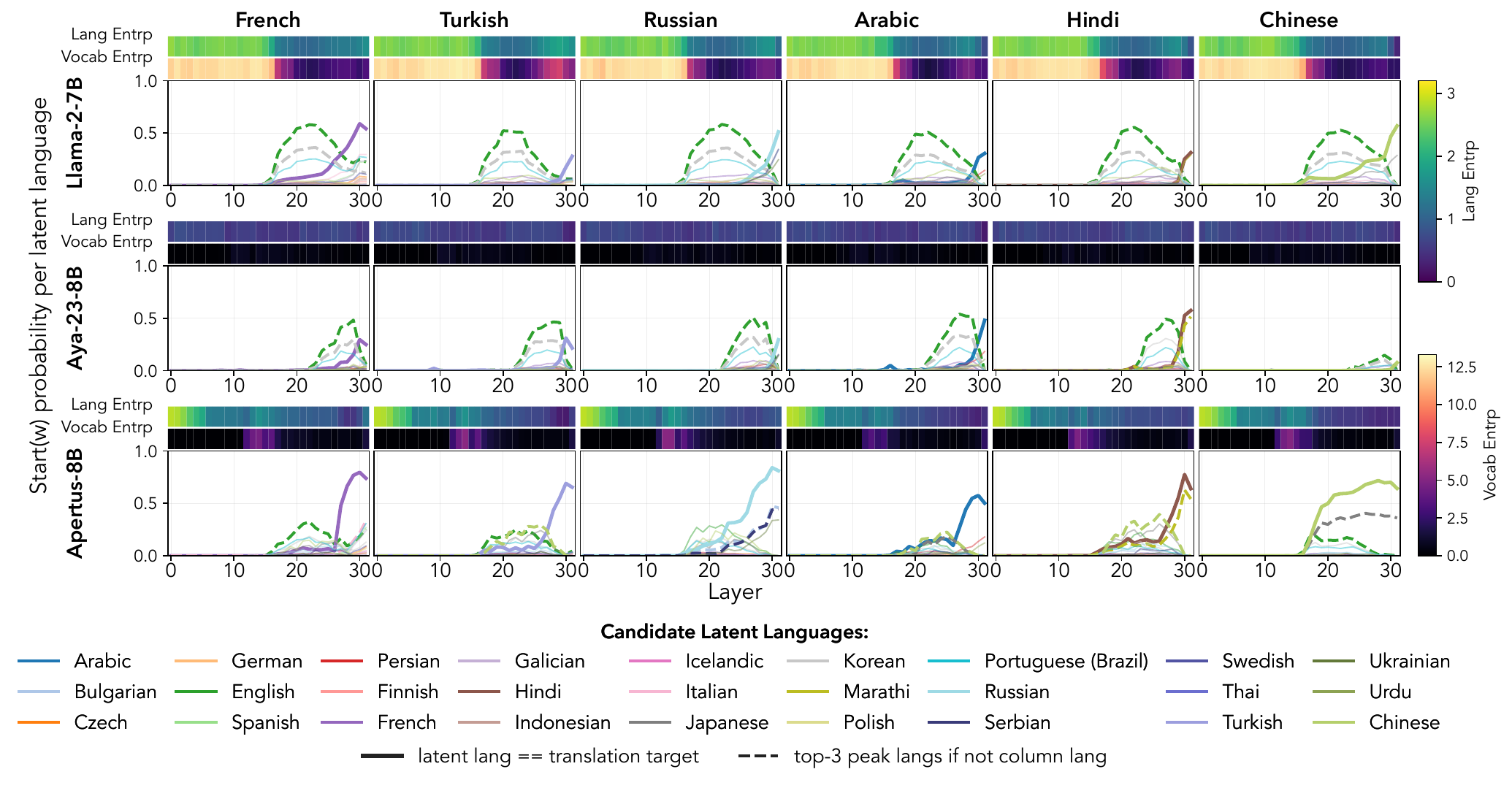}
    \caption{\textbf{Latent language probability rises in later layers in controlled translation; English is a stronger competing latent language in Llama-2 than in Aya-23 and Apertus.} For each translation prompt and model layer, we measure the summed next-token probability of all tokens that can begin the corresponding answer in each candidate language ($\startw{}$ probability). Rows are models and columns group prompts with the same translation targets. English is excluded as a source or target, but retained as a candidate latent language. Upper strips show average latent language distribution entropy and average first-token vocabulary entropy.}
    \label{fig:translation-startw-by-target-no-en}
    \vspace{-1em}
\end{figure*}

\subsection{LLID Estimators}
For open-ended evaluation, all estimators produce a layerwise distribution $q$ over candidate languages. We compare:
\begin{itemize}[
      leftmargin=1em,
      itemsep=0pt,
      topsep=0pt,
      parsep=0pt,
      partopsep=0pt
    ]
    \item \repr: a representation-based GMM posterior fit to multilingual hidden states;
    \item \decargmax: decoding-based LLID using deterministic argmax rollouts from intermediate layers;
    \item \dectopp: decoding-based LLID using top-$p$ rollouts.
\end{itemize}
For \decargmax{} and \dectopp{}, we generate short continuations from intermediate layers, pass them through GlotLID \citep{kargaran2023glotlid}, and aggregate the resulting per-language scores into $q$. We evaluate both the raw logitlens and tuned lens defined in \S\ref{sec:framework-evidence}, separating the rollout procedure from the projection used to map intermediate hidden states into the output vocabulary. Fitting details for GMMs, tuned lenses, and GlotLID are in Appendix~\ref{sec:appendix_training}.

\begin{figure*}[th]
    \centering
    \includegraphics[width=\textwidth]{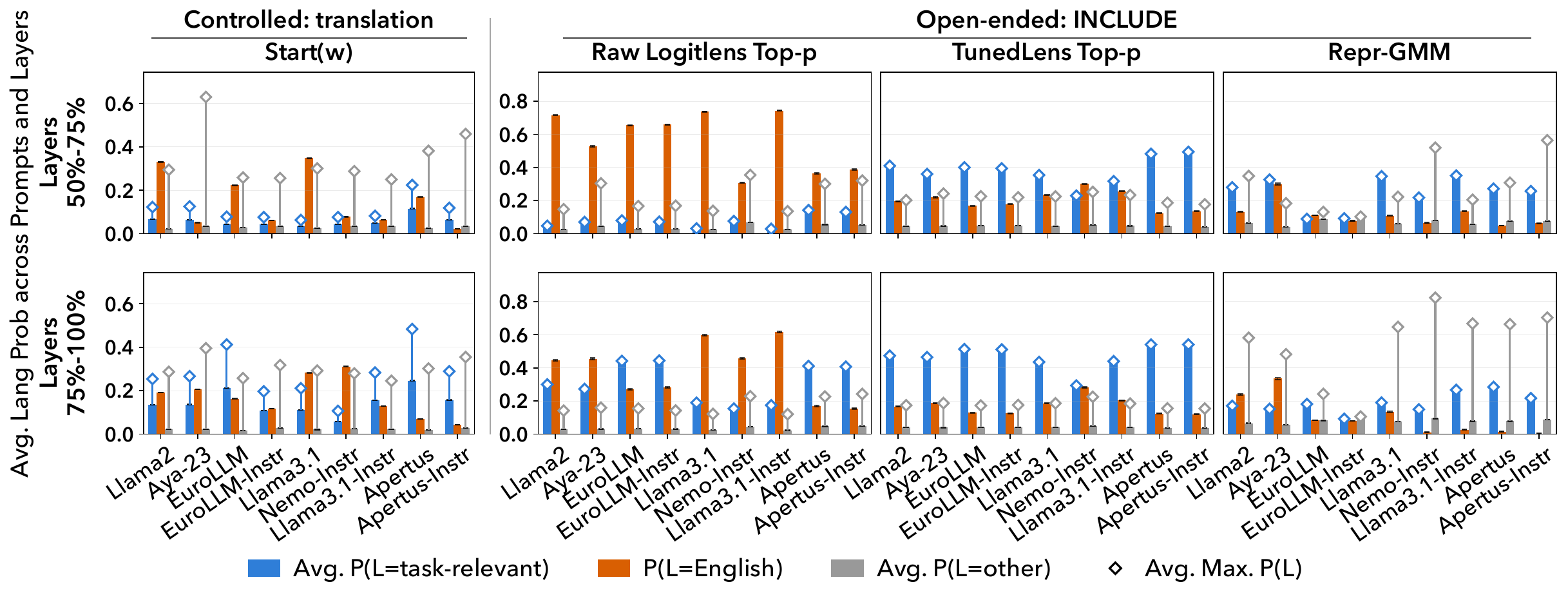}
    \caption{\textbf{Later-layer latent language estimates depend strongly on the probe: raw logitlens assigns more probability to English, while tuned lens and GMM estimates favor task-relevant languages.} Rows aggregate the 50--75\% and 75--100\% layer windows. \textbf{Left:} translation prompts using normalized $\startw{}$ probabilities. \textbf{Right:} open-ended INCLUDE prompts using raw logitlens top-$p$, tuned lens top-$p$, and representation-based GMM LLID. Bars pool prompt--layer--language probabilities within each category before averaging; diamonds show the average maximum probability within the task-relevant or other-language category. Error bars show SE. Cases with English as source or target language are excluded.}
    \label{fig:language-probs-late-layer-windows}
\end{figure*}
 
\section{The Controlled Case: Latent Language Under Known Completions}
\label{sec:synthetic-reproduced}

We begin by testing decoding-based LLID in the controlled setting, where the set of valid completions is known across languages and we can directly track when language-specific completions become decodable from intermediate layers.
We track the summed $\startw{}$ probability assigned to language-specific tokens of the same concept across layers for three models ordered by increasing multilingual capability: Llama-2-7B, Aya-23-8B, and Apertus-8B. Consistent with prior observations on Llama-2-7B, we find that across all model families the decoded language signal is weak in early layers and becomes most visible in later layers across translation (Fig.~\ref{fig:translation-startw-by-target-no-en}), copy (Fig.~\ref{fig:app-copy-startw}), and cloze tasks (Fig.~\ref{fig:app-cloze-startw}). However, the sharpness of this signal differs substantially across models with different degrees of multilinguality. The effect is also not necessarily uniform across tasks and languages within a model.

Across models, the copy task yields sharper and higher late-layer probabilities for the observed language. This is expected: copying can rely on reproducing a surface form that is already present in the context. Cloze, on the other hand, requires retrieving a concept and lexicalizing it in the appropriate language\footnote{Expressing an underlying concept using the words/forms of a particular language.}. Compared with copy, the cloze plots show lower peaks, more competing languages, and more diffuse probability mass, especially for Llama-2 and Aya-23.

Even when English is excluded from the translation direction (Fig.~\ref{fig:translation-startw-by-target-no-en}), Llama-2 frequently maintains relatively high probability mass on English lexicalizations across layers, suggesting a stronger English-centered bias in its internal representations. This effect is weaker in Aya and further reduced in Apertus, where the target language probability rises relatively earlier and becomes more consistently dominant in later layers.

Taken together, these results show that $\startw{}$ recovers a meaningful late-layer language signal in the controlled setting. The diagnostic is strongest when the task makes the relevant surface form directly available, as in copy prompts, and weaker when the model must retrieve and lexicalize a concept, as in cloze prompts. The diagnosis also varies systematically with model multilinguality: less multilingual models more often retain high-probability English alternatives, while the more multilingual Apertus-8B shifts probability toward the intended non-English target earlier.

\begin{figure*}[th]
  \centering
  \includegraphics[width=\textwidth]{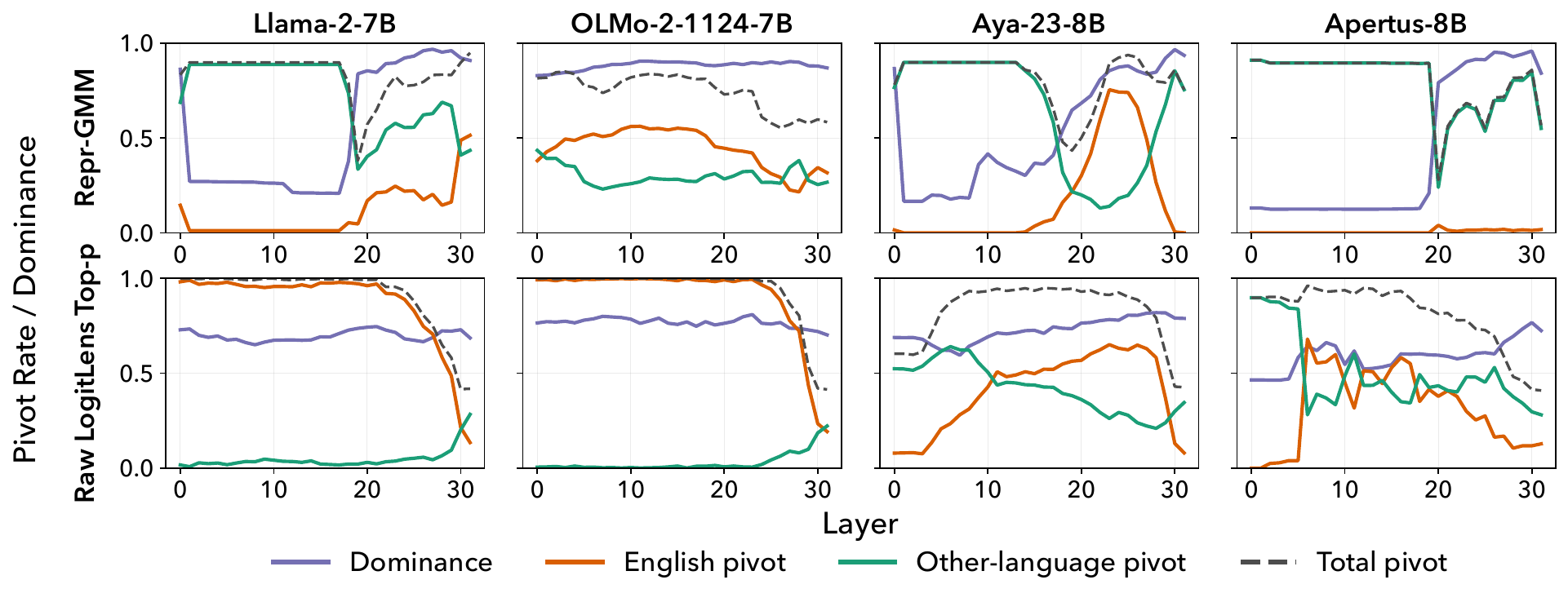}
  \caption{\textbf{In open-ended INCLUDE, latent language dominance and pivoting vary across models and probes.} Representation-based estimates exhibit model-specific changes across the layer stack, whereas decoding-based estimates generally shift later, depending on model multilinguality. For each prompt and layer, dominance is the highest probability assigned to a candidate latent language; a pivot occurs when the dominant language differs from the prompt language. Rows are LLID estimators and columns are models. Pivot rates are separated into English and other languages; the dashed line shows their sum. INCLUDE contains no English prompts.}
  \label{fig:include-pivot-cat-dominance}
\end{figure*}

\section{Generalizing Decoding-based Estimates to Open-Ended Generation}
\label{sec:open-ended}

The controlled setting above provides a favorable testbed for the $\startw{}$ measurement as the relevant cross-lingual completions are known in advance, so the diagnostic can compare a fixed set of lexical alternatives. However, for $\startw{}$ to support claims about latent language or multilingual pivoting, its signal should generalize beyond controlled settings with predefined candidate answers. We therefore ask whether similar language preferences can be recovered in open-ended next-token prediction, where the model generates continuations in context rather than selecting among predefined translations.

We extend the decoding diagnostic to open-ended generation by generating short continuations from intermediate hidden states under either the raw logitlens or tuned lens, using argmax or top-$p$ decoding, and then classifying the continuation language.
Fig.~\ref{fig:language-probs-late-layer-windows} shows that the projection method changes how probability mass is allocated across languages. Raw logitlens decoding assigns substantially more mass to English than the controlled $\startw{}$ setting, suggesting that English lexicalizations remain accessible when the model is not restricted to known translations. For some models, raw logitlens outputs also become difficult to interpret, given the representational drift between intermediate and final-layer hidden states \citep{belrose2025tunedlens}. The tuned lens shifts average probability toward the task-relevant language while average probability assigned to other languages remains low. This suggests that the tuned lens may partly collapse intermediate language evidence toward the final decoded language, consistent with the concern raised by \citet{wendler2024llamas}.

\begin{figure*}[th]
    \centering
    \includegraphics[width=\textwidth]{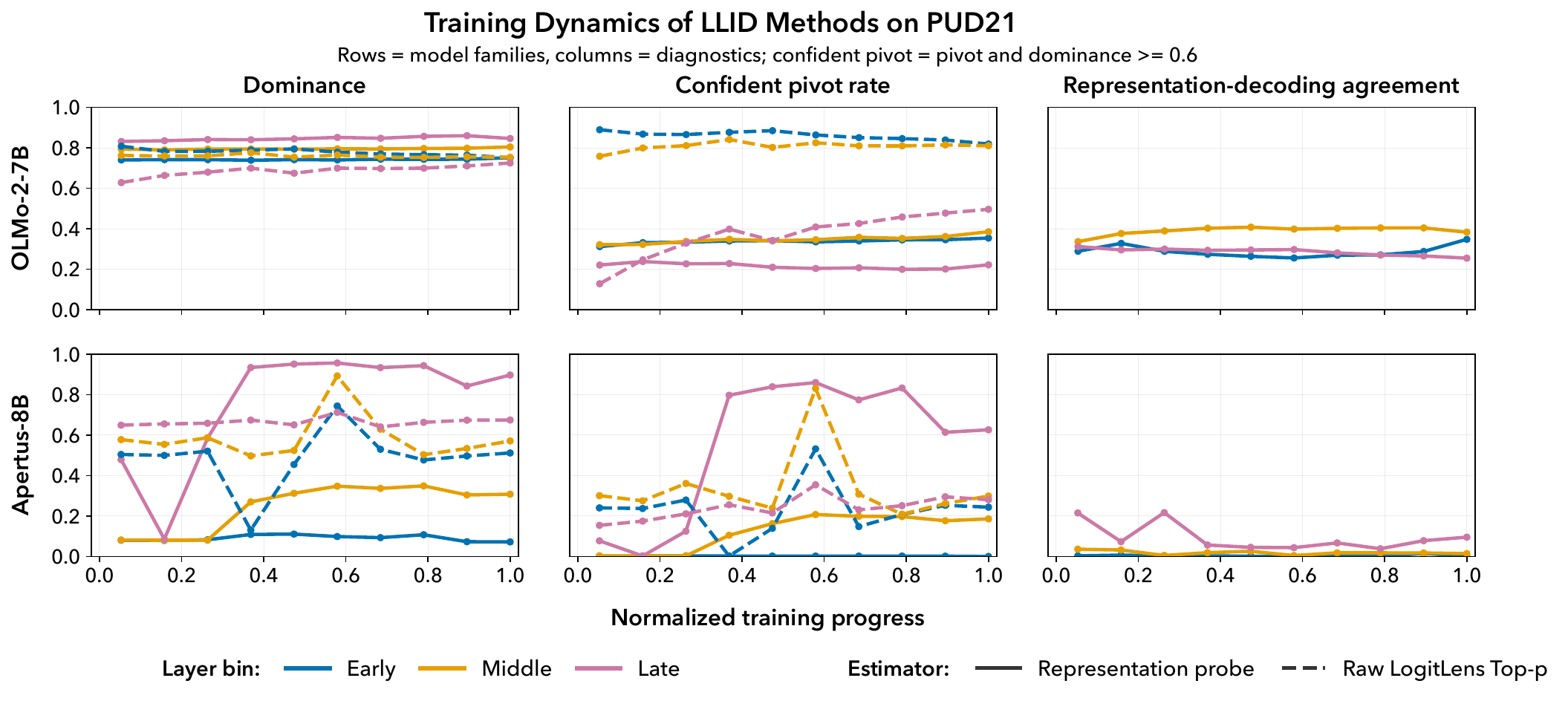}
    \caption{\textbf{Training dynamics of latent language estimates across pretraining on PUD21.} 
    Rows are model families and columns are metrics. 
    Colors denote early, middle, and late layer bins; solid and dashed lines denote \repr{} and \dectopp{}, respectively.
    Dominance is the average top-language probability; confident pivot rate is the fraction of prompts where a non-task language is dominant with probability at least 0.6; agreement measures whether the two probes select the same dominant language. The $x$-axis is normalized training progress, spanning ${\sim}$210B to 4T training tokens. Layer bins use normalized layer index $j/J$: early $j/J<0.25$, middle $0.25\leq j/J<0.75$, and late $j/J\geq0.75$. Most checkpoint-dependent changes occur in late layers, while representation--decoding agreement remains low across training.}
    \label{fig:training-dynamics}
\end{figure*}

\section{Representation vs.\ Decoding: Different Evidence, Different LLID Behaviors}
\label{sec:repr-vs-dec}

Having established how decoding-based estimates vary across projections, rollouts, and task settings, we now compare them with representation-based LLID on the same hidden states. 
Representation-based LLID provides a complementary view of the same intermediate states examined by decoding-based probes. Instead of asking what language can be decoded from a hidden state, it asks which language-conditioned region of activation space the hidden state resembles. This distinction matters because a hidden state may be geometrically closer to one language's cluster without being decodable as that language, and conversely, an intermediate state may decode into a language whose representation-space signature is not dominant. Disagreement between the two estimators should therefore be treated as measurement information rather than as immediate diagnostic failure. Such disagreement provides evidence that the probes capture different aspects of multilingual processing.

Comparing the rightmost column of Fig.~\ref{fig:language-probs-late-layer-windows} (Repr-GMM) with the decoding-based columns on the same open-ended INCLUDE prompts makes this contrast visible. The representation probe assigns substantially more probability mass to the task-relevant language and correspondingly less to English than raw logitlens top-$p$ decoding does, particularly in the 50--75\% layer window. This gap is most pronounced for models with moderate multilinguality (\eg{}, Llama-3.1), where decoding-based estimates still show a strong English bar while the representation-based estimate already favors the task languages. For the most multilingual models (\eg{}, Apertus, Apertus-Instruct), the average probabilities suggest closer agreement as both probes assign relatively little probability to English and substantial probability to the task-relevant language in late layers. However, the maximum-language estimates also reveal that, under the representation probe, a single non-task, non-English language can still receive more probability than the task-relevant language. These patterns suggest that the two estimator families are sensitive to different aspects of the same hidden states. The GMM representation probe reflects the geometric neighborhood of the input language's activation, while decoding-based estimators reflect the accessibility of language-specific tokens through the unembedding matrix, which retains a stronger English bias even when the representation probe assigns substantially less probability to English.

The precise layerwise patterns are also model-dependent (Fig.~\ref{fig:include-pivot-cat-dominance}). For the representation probe, Llama-2 exhibits a mixture of English- and other-language pivots, Aya-23 shows a pronounced but temporary increase in English-directed pivots, while Apertus-8B pivots are almost entirely to non-English languages. OLMo-2 shows another distinct pattern, maintaining high representation-probe dominance across nearly the entire layer stack. By contrast, dominance drops sharply near the start and recovers through the middle-to-late layers for Llama-2 and Aya-23, while Apertus-8B remains low until a late-layer rise.

These patterns reflect again the different evidence used by the two probe families. The representation probe is fit directly to input-language activation distributions, so it is most confident where hidden states preserve a strong language-conditioned geometric signature. As states become more mixed or task-oriented, they are less cleanly assigned to a single fitted language distribution, producing the early dominance drop in Llama-2 and Aya-23 and the low early-to-middle dominance in Apertus. Their later recovery indicates that the states become concentrated again in a language-conditioned region. OLMo-2 is the exception, as its high dominance suggests a persistent separation of language-conditioned activation regions rather than a pronounced early-to-mid layer mixing phase. Decoding-based estimators, on the other hand, generally change much later in the network.

\paragraph{Across Domains.}
\label{sec:domain-specific}
We next ask whether the model and estimator-specific trajectories in Fig.~\ref{fig:include-pivot-cat-dominance} persist across language inventories and QA subject domains. Fig.~\ref{fig:app-domain-dominance-pivot} repeats the dominance and pivot-rate analyses across PUD language-set variants and INCLUDE domains. For a given model and LLID estimator, the curves largely preserve the same layerwise shape across settings. Thus, domain and language inventory changes affect absolute levels less than model or diagnostic choice. The clearest systematic exception is candidate-inventory size: PUD21 generally yields lower dominance and higher pivot rates than PUD9, as expected when probability mass is distributed over more possible languages.

\paragraph{Across Pretraining.}
\label{sec:training-dynamics}
We ask whether latent language estimates appear suddenly, consolidate gradually, or drift as the model trains. Fig.~\ref{fig:training-dynamics} compares OLMo-2-7B and Apertus-8B across checkpoints (see Table~\ref{tab:olmo-apertus-token-aligned-checkpoints} for checkpoint alignment), tracking representation and decoding dominance, confident pivot rate, and estimator agreement in early, middle, and late layer bins.
We use confident pivot rate to avoid treating small fluctuations in diffuse LLID distributions as meaningful pivots, retaining only cases where a non-task language is dominant with probability at least 0.6.
The two models show qualitatively different trajectories. For OLMo, representation-based dominance is high and stable from the earliest checkpoint, while decoding dominance consolidates over the first ${\sim}$20--30\% of training; confident pivoting is low for \repr{} but moderate and slowly increasing for \dectopp{}, consistent with English-centric training where the representation space separates languages early but the decoding pathway retains an English bias. For Apertus, both estimators undergo a sharp transition between roughly 20\% and 40\% of training, after which late-layer dominance reaches ${\sim}$0.9. Representation--decoding agreement remains low throughout training for both models, and is especially low for Apertus. These dynamics suggest that most of the checkpoint-dependent changes occur in the late-layer bin for both models.
Overall, the checkpoint analysis shows that representation- and decoding-based estimates of LLID behavior can change at different points in training.

\paragraph{Base vs.\ Instruct.}
\label{sec:base-instruct}
Finally, we ask whether instruction tuning increases English dominance in LLID estimates. This hypothesis is plausible because instruction data is often less multilingual than pretraining data, so instruction tuning might reorient intermediate states or decoded continuations toward English. In Fig.~\ref{fig:language-probs-late-layer-windows}, base and instruction-tuned variants show only modest differences in their average LLID probabilities. The layerwise comparisons in the Appendix (Figs.~\ref{fig:app-base-vs-instruct-pud}--\ref{fig:app-base-vs-instruct-include}) likewise show broadly similar pivot and entropy trajectories, although some model-specific deviations remain. Contrary to our initial expectation, instruction tuning produces little systematic shift toward English in LLID estimates, and its overall effects are considerably smaller than the changes observed across pretraining checkpoints.
 
\section{Related Work}

\paragraph{Shared and language-specific multilingual \mbox{structure}.}
Multilingual models maintain both shared and language-specific structure in hidden states, through related but distinguishable subspaces \citep{chang-etal-2022-geometry}, language-neutral subnetworks \citep{foroutan2022discovering}, and language-specific neurons or components \citep{zeng2025linguafranca,zhao2025languagereasoningdisentanglement}. These findings provide a basis for representation-based LLID, while highlighting that hidden states can simultaneously encode shared and language-specific structure.

\paragraph{Latent pivots and output language.}
A second line of work studies whether multilingual models internally route computation through a pivot language. Decoding-based probes find English-pivot behavior in English-dominated models and different pivoting patterns under non-English-centric training \citep{wendler2024llamas,zhong2025language}. Causal interventions further separate output language from conceptual content, suggesting that language form and meaning can be partially disentangled in the residual stream \citep{dumas2025separatingtonguefromthought}. Our work differs by asking whether the probes used to infer such latent language behavior agree on the same hidden states.

\paragraph{Training dynamics of multilingual representations.}
Recent work studies how multilingual abilities emerge during training rather than only at the final checkpoint, showing that cross-lingual abilities \citep{blevins2022crosslingual} and internal linguistic features \citep{bayazit2026crosscoding} develop, stabilize, or change over pretraining. Accordingly, we examine LLID across training checkpoints: if latent language estimates reflect model properties rather than diagnostic artifacts, they should vary systematically with training progression.

\section{Conclusion}
We reframed latent language identification as a measurement problem, comparing representation-based and decoding-based probes across models, training regimes, and tasks. The GMM representation probe and decoding-based probes disagree systematically: the former identifies language structure earlier and with weaker English bias, while decoding pathways retain sharper, more English-favored signals. These differences vary systematically with multilinguality and training progression, but their exact layerwise trajectories remain model and lens-dependent and cannot be predicted from multilinguality alone. Domain and language-inventory changes, by contrast, shift LLID estimates only modestly. Rather than confirming a single internal \textit{lingua franca}, existing probes expose complementary aspects of multilingual computation, showing that the choice of probe is itself part of the claim.

\section*{Acknowledgements}
We thank Clara Meister, Negar Foroutan, Angelika Romanou, and Ayush K. Tarun for their helpful discussions and feedback on our manuscript.
We also gratefully acknowledge the support of the Swiss National Science Foundation (No. 215390), the AI2050 program at Schmidt Sciences (Grant \#G-25-69783), Sony Group Corporation, and the Swiss National Supercomputing Center (CSCS) in the form of an infrastructure engineering and development project. This research is with support from Google.org and the Google Cloud Research Credits program for the Gemini Academic Program. This project is also funded by the European Union (ERC, RESPECT-LM, 101222478). Views and opinions expressed are however those of the author(s) only and do not necessarily reflect those of the European Union or the European Research Council. Neither the European Union nor the granting authority can be held responsible for them.

\section*{Limitations}

Our findings come with several limitations. We evaluate decoder-only models in the 7--9B parameter range, and our checkpoint analysis is necessarily restricted to OLMo-2 and Apertus, the two model families that release intermediate training checkpoints publicly. Whether probe disagreement persists at other scales, or for non-decoder architectures, remains an open question. The precise layerwise LLID patterns also vary across model families, so the trajectories observed here should not be assumed to generalize unchanged to other models.
We compare only two probe families, of which the representation-based family is instantiated only with the GMM estimator \citep{shani2025language}, and do not examine causal interventions \citep{dumas2025separatingtonguefromthought} or neuron-level attribution \citep{zeng2025linguafranca}, which may align with either family or expose a third aspect of multilingual processing. Because latent language is inferred rather than observed (\ie{}, no ground truth), our work documents probe disagreement but cannot adjudicate which (if any) best reflects internal computation. Probe-specific choices (\eg{}, GMM components, the tuned lens fitting corpus, the open-ended anchor position, and the concept-word construction) may shift absolute LLID values, though we expect qualitative trends to be robust.
Finally, our 27-language inventory excludes very low-resource languages and formal languages such as code and mathematics. We also leave open whether probe disagreement predicts downstream failure modes in cross-lingual transfer or generation.

\section*{Ethics Statement}
This work uses publicly available pretrained language models and multilingual evaluation datasets (PUD/UD, INCLUDE, and Fineweb2), used in accordance with their respective licenses. Our analysis is diagnostic in nature and does not involve training new models, collecting human data, or deploying systems in user-facing settings. We see no direct ethical concerns arising from this work.

\bibliography{custom}

\appendix
\begin{table*}[th]
\centering
\small
\resizebox{1.0\linewidth}{!}{
    \begin{tabular}
    {
        >{\raggedright\arraybackslash}p{0.12\linewidth}
        >{\raggedleft\arraybackslash}p{0.07\linewidth}
        >{\raggedleft\arraybackslash}p{0.22\linewidth}
        >{\raggedright\arraybackslash}p{0.48\linewidth}
    }
    \toprule
    \textbf{Setting} & \textbf{\# Langs} & \textbf{\# Entries} & \textbf{Languages} \\
    \midrule
    PUD9 & 9 & 900 train + 900 held-out & Arabic (ar), Czech (cs), English (en), French (fr), Hindi (hi), Icelandic (is), Indonesian (id), Portuguese (pt), Spanish (es) \\
    PUD21 & 21 & 2,100 train + 2,100 held-out & Arabic (ar), Russian (ru), Hindi (hi), Chinese (zh), Korean (ko), Japanese (ja), Indonesian (id), German (de), English (en), Icelandic (is), Swedish (sv), Spanish (es), French (fr), Galician (gl), Italian (it), Portuguese (pt), Czech (cs), Polish (pl), Turkish (tr), Finnish (fi), Thai (th) \\
    UD6 extension & 6 & 600 train + 600 held-out & Ukrainian (uk), Bulgarian (bg), Serbian (sr), Urdu (ur), \hspace{5em} Persian (fa), Marathi (mr) \\
    INCLUDE-10 & 10 & 882 prompts & Arabic (90), Spanish (90), Finnish (90), French (81), Hindi (90), Indonesian (90), Portuguese (90), Russian (90), \hspace{5em} Turkish (90), Chinese (81) \\
    Copy/cloze & 6 & 414 prompts per task & Arabic (ar), Hindi (hi), Chinese (zh), Russian (ru), \hspace{5em} English (en), French (fr) \\
    Translation & 9 & 16,146 prompts & Arabic (ar), Spanish (es), Finnish (fi), French (fr), Hindi (hi), Indonesian (id), Russian (ru), Turkish (tr), Chinese (zh) \\
    \bottomrule
    \end{tabular}
}
\caption{\textbf{Language inventories used in the experiments.} PUD and UD rows are balanced at 100 train and 100 held-out prompts per language. PUD9+UD6 and PUD21+UD6 add the UD6 extension to the corresponding PUD inventory. The INCLUDE subset is capped at 30 prompts per language-domain cell, but French and Chinese have fewer total prompts because their STEM cells contain only 21 selected examples. Target-string copy and cloze counts are for the six-language main-plot subset; translation counts aggregate the nine target-specific runs.}
\label{tab:dataset-language-inventories}
\end{table*}

\newpage

\section{Dataset Details}
\label{sec:appendix_dataset}

In this section we describe the datasets and the preprocessing steps. Note that we did not collect these datasets ourselves; they may contain personally identifying or offensive content.

\subsection{Language Inventory}
Table~\ref{tab:dataset-language-inventories} gives the language inventories used across the controlled target-string runs, PUD runs, and INCLUDE runs. 
In the main text, we use the shorthand PUD9, PUD21, PUD9+UD6, PUD21+UD6 to avoid repeatedly listing these languages. Entry counts refer to prepared prompt rows; for PUD and UD settings, we report the deterministic train and held-out splits separately. The PUD9+UD6 and PUD21+UD6 settings are formed by adding the UD6 extension row to PUD9 or PUD21.

\subsection{Controlled Target-String Construction} \label{sec:appendix_target_string_construction}
The controlled copy, cloze, and translation prompts are derived from the target-word setup of prior latent language studies \citep{wendler2024llamas,dumas2025separatingtonguefromthought,zhong2025language}. Each example has a known concept completion in multiple languages, and the logitlens query is made at the final task token, before the answer begins. For instruction-tuned models, formatting may append assistant-control tokens, so we anchor scoring to the end of the original task prefix rather than the final formatted token.

We first build a common-concept grid and then derive copy, cloze, and directed translation prompts from that grid. Copy and translation prompts use four completed in-context examples followed by a query line; cloze prompts use two completed cloze demonstrations followed by the query context. For translation and cloze settings, word translations and cloze contexts are constructed jointly so that the context disambiguates homonyms. We filter examples where the translated cloze answer does not align with the independently translated concept word. We do not remove examples solely because the English and target language forms share a tokenizer prefix.

For each remaining prompt $x$, candidate language $\ell$, and model tokenizer $\tau$, we construct $\startw$, the set of vocabulary tokens that can begin the language-$\ell$ answer $w_{\ell,x}$, as defined in \S\ref{sec:eval-regimes}. 
This set includes vocabulary tokens with tokenizer-specific prefixes and whitespace variants when applicable. For tokenizers with byte-fallback behavior, we also include valid byte-level starting tokens. 

The corresponding score $s_x(\ell)$, also defined in \S\ref{sec:eval-regimes}, is the summed next-token probability assigned to these valid starting tokens. We use this raw summed mass for the probability curves. When computing distributional metrics such as entropy, dominance, or pivot rate, we normalize these scores over the candidate language set:
\[
q_x(\ell)
=
\frac{s_x(\ell)}
{\sum_{\ell'\in\langs}s_x(\ell')}
\]
The language entropy is computed from this normalized distribution $q_x$, while vocabulary entropy is computed over the model's next-token vocabulary distribution.

\subsection{Open-Ended Evaluation Details}
\label{sec:appendix_openended_eval}
For natural-language data without constrained completions, both representation-based and decoding-based LLID are evaluated at a matched prompt anchor. Unless otherwise stated, we use the midpoint of the prompt and aggregate over a forward window of five positions.
Decoding-based open-ended LLID uses either deterministic argmax rollouts or top-$p$ rollouts with $p=0.9$ and five samples. We score each decoded continuation with GlotLID \citep{kargaran2023glotlid}, collapse its labels to the candidate-language inventory, and normalize the resulting distribution. For top-$p$ rollouts, we average these distributions uniformly across samples to obtain $q$. We use PUD as sentence-level natural-language data. PUD21 is the broadest setting, while PUD9 follows the language subset used by \citet{shani2025language}. We also evaluate PUD9+UD6 and PUD21+UD6 by adding six Universal Dependencies languages selected for script or family overlap. For domain analyses, we use INCLUDE question-answering prompts in 10 languages and three domains: Social Science, Arts \& Humanities, and STEM.

\section{Model Details} 
\label{sec:appendix_model}

Table~\ref{tab:model-stats} reports all evaluated models and multilinguality proxies. Since direct pretraining mixtures are not fully available, we use zero-shot QA accuracy on INCLUDE, cross-lingual perplexity, and byte-normalized likelihood as external proxies for multilingual capability. For the training dynamics analysis, we align OLMo-2 and Apertus checkpoints as shown in Table~\ref{tab:olmo-apertus-token-aligned-checkpoints}.

The INCLUDE score is zero-shot accuracy from \texttt{lm-evaluation-harness} runs on \texttt{include-base-44}, macro-averaged over 44 language groups. We use it to sort the models because it directly measures multilingual task performance.

The PUD21 line NLL is a likelihood-based auxiliary score on the PUD21 parallel sentence set. We compute the autoregressive negative log-likelihood of each teacher-forced line, convert it to bits, and average across PUD-only parallel rows in the 21 languages. Lower is better.

The PUD21+UD6 BPB score is a byte-normalized likelihood on the broader PUD21+UD6 corpus. We sum token-level negative log-likelihood over all evaluated lines, convert from nats to bits, and divide by the total number of UTF-8 bytes:
\[
\mathrm{BPB}
=
\frac{\sum_x -\log_2 p_\theta(x)}
{\sum_x |\mathrm{utf8}(x)|}
\]
This reduces, but does not eliminate, tokenization and script effects; lower BPB is better.

\begin{table}
\centering
\resizebox{1.0\linewidth}{!}{
    \begin{tabular}{lccc}
        \toprule
        \textbf{Model} &
        \multicolumn{1}{c}{\makecell{\textbf{INCLUDE}\\\textbf{0-shot macro} ($\uparrow$)}} &
        \multicolumn{1}{c}{\makecell{\textbf{PUD21 line}\\\textbf{NLL} ($\downarrow$)}} &
        \multicolumn{1}{c}{\makecell{\textbf{PUD21+UD6}\\\textbf{BPB} ($\downarrow$)}} \\
        \midrule
        Apertus-8B-Instruct & \textbf{55.0} & 169.8 & 1.19 \\
        Apertus-8B \citep{apertus2025apertus} & \underline{53.1} & \textbf{138.3} & \textbf{0.97} \\
        Llama-3.1-8B-Instruct & 52.8 & 152.9 & 1.09 \\
        Mistral-Nemo-Instruct-2407 \citep{mistralai2024mistralnemo} & 52.2 & 166.7 & 1.20 \\
        Llama-3.1-8B \citep{grattafiori2024llama3herdmodels} & 48.9 & \underline{148.8} & \underline{1.05} \\
        EuroLLM-9B-Instruct & 47.8 & 156.0 & 1.17 \\
        EuroLLM-9B \citep{martins2025eurollm} & 43.3 & 149.2 & 1.11 \\
        Aya-23-8B \citep{aryabumi2024aya23} & 40.1 & 172.0 & 1.26 \\
        OLMo-2-1124-7B \citep{walsh2025olmo2} & 33.8 & 181.1 & 1.30 \\
        Llama-2-7B \citep{touvron2023llama2} & 27.7 & 170.0 & 1.22 \\
        GPT-2 \citep{radford2019gpt2} & 26.2 & 365.4 & 2.62 \\
        GPT-2-XL & 25.9 & 302.3 & 2.20 \\
        \bottomrule
    \end{tabular}
}
\caption{\textbf{Multilingual Performance.}
Models are sorted by decreasing zero-shot macro accuracy on INCLUDE. 
Line NLL is an auxiliary likelihood score computed on a parallel corpus across languages.
BPB is likelihood normalized by UTF-8 byte count, but it remains script-sensitive because byte counts differ across writing systems.
Best is bold; second-best is underlined.}
\label{tab:model-stats}
\end{table}

\begin{table}[t]
\centering
\small
\resizebox{0.8\linewidth}{!}{
    \begin{tabular}{ccc}
    \toprule
    \textbf{Alignment Idx} & \textbf{OLMo-2 tokens} & \textbf{Apertus tokens} \\
    \midrule
    1  & 210B  & 210B \\
    2  & 625B  & 630B \\
    3  & 1.04T & 1.05T \\
    4  & 1.46T & 1.47T \\
    5  & 1.87T & 1.89T \\
    6  & 2.29T & 2.31T \\
    7  & 2.70T & 2.73T \\
    8  & 3.12T & 3.15T \\
    9  & 3.53T & 3.57T \\
    10 & 3.95T & 3.99T \\
    \bottomrule
    \end{tabular}
}
\caption{\textbf{Token-aligned checkpoints,} used in the OLMo-2 and Apertus training dynamics comparison.}
\label{tab:olmo-apertus-token-aligned-checkpoints}
\end{table}

\begin{figure*}[th]
    \centering
    \includegraphics[width=\textwidth]{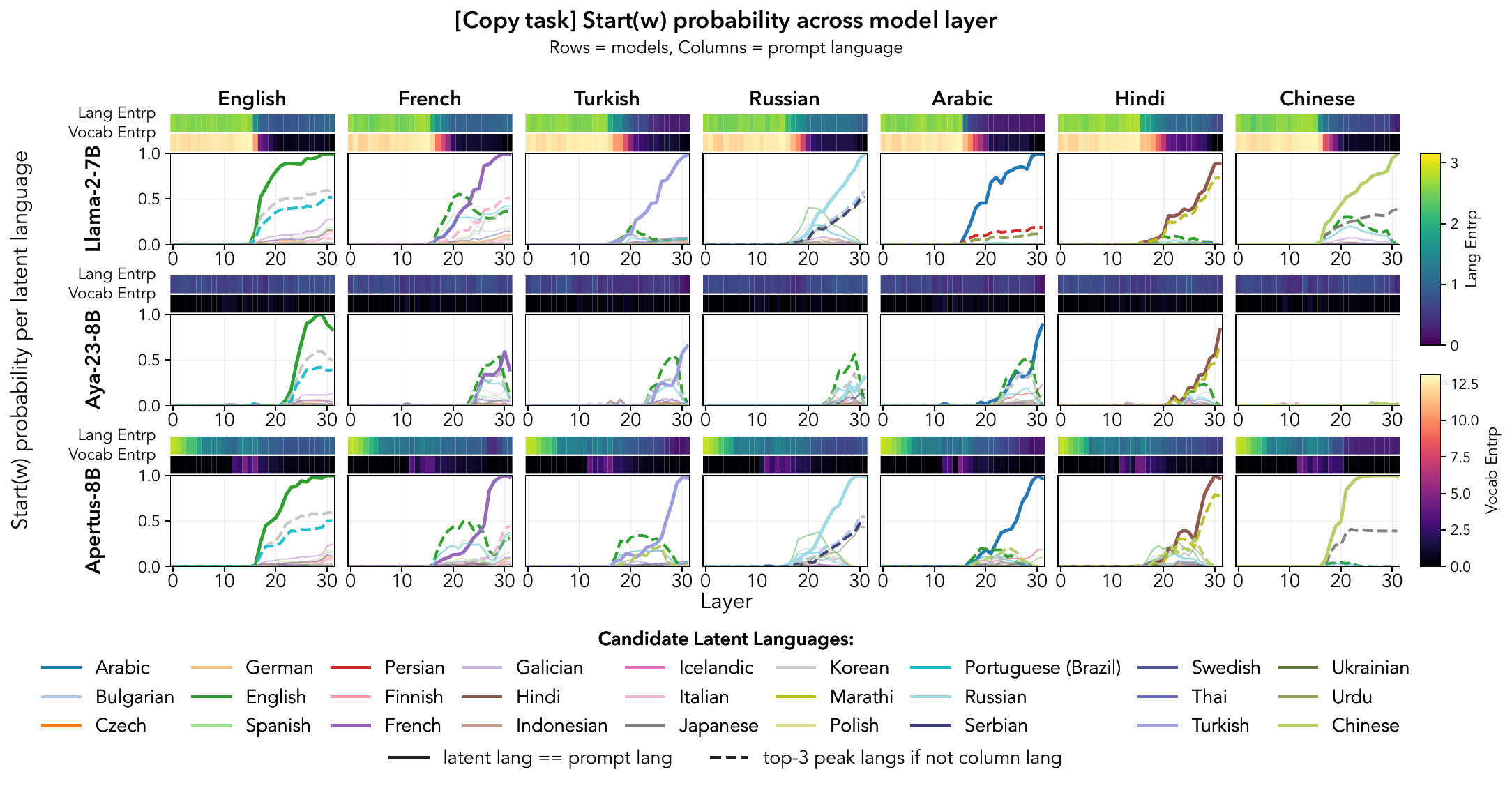}
    \caption{\textbf{Copy prompts: latent language probability across model layers.} For each copy prompt and model layer, we measure the summed next-token probability of all tokens that can begin the corresponding answer in each candidate language ($\startw{}$ probability). Rows are models and columns group prompts with the same prompt language. Upper strips show average latent language distribution entropy and average first-token vocabulary entropy.}
    \label{fig:app-copy-startw}
\end{figure*}

\section{Training Details}
\label{sec:appendix_training}

\paragraph{Hardware, Packages \& Artifacts.}
We run experiments on NVIDIA A100 80GB GPUs, using one GPU at a time. Each run lasts between one to six hours depending on the dataset size. We use Python 3.10 and use transformers\footnote{\url{https://huggingface.co/docs/transformers}}, nnsight\footnote{\url{https://nnsight.net/}}, and datasets\footnote{\url{https://huggingface.co/datasets}} packages. We used AI-assisted tools to support code development, and we reviewed and tested the resulting code. For writing, AI tools were used to rephrase the wording of the manuscript.

\paragraph{GMM Fitting.}
Representation-based LLID requires a language-conditioned model of activation space. We fit one per-layer Gaussian model per candidate language from hidden states collected on the train split. Each GMM setup uses at most 100 training prompts per language, uniform language priors, layerwise PCA retaining 98\% explained variance, and diagonal covariance. Unless otherwise stated, we fit token-level hidden states and use the surface-token alignment from the dataset preprocessing to select comparable positions at evaluation time.
We fit separate GMMs per candidate-language inventory: \texttt{pud9}, \texttt{pud21}, \texttt{pud9\_ud6}, \texttt{pud21\_ud6}, and \texttt{include\_10lang}. For INCLUDE plots that contain English as a possible latent language, we fit with the same PUD calibration source but add English to the candidate-language set.

\paragraph{Tuned Lens Fitting.}
For decoding-based LLID, we compare raw logitlens projections with tuned lens projections \citep{belrose2025tunedlens}. Tuned lenses are fit on multilingual FineWeb2 data \citep{penedo2025fineweb2}, keeping the fitting corpus independent of PUD, INCLUDE, and controlled target-string evaluation prompts. Our main tuned lens artifacts use the 27-language inventory.
For each model, FineWeb2 rows are tokenized into windows of length up to 2048 tokens, subject to the model's supported context length. Training examples are balanced by strict round-robin over languages, dropping excess rows from higher-resource languages. We train for one epoch with at most 1000 optimization steps, AdamW with learning rate $10^{-3}$ and weight decay 0, temperature 1.0, identity and bias regularization weights $10^{-4}$, and bfloat16 autocast. Validation uses a lightweight balanced subset, and checkpoints are written periodically. We do not train a tuned lens translator for the final transformer layer; at runtime, the final layer uses the raw model-head projection.

\section{Additional Results}
\begin{figure*}[th]
    \centering
    \includegraphics[width=\textwidth]{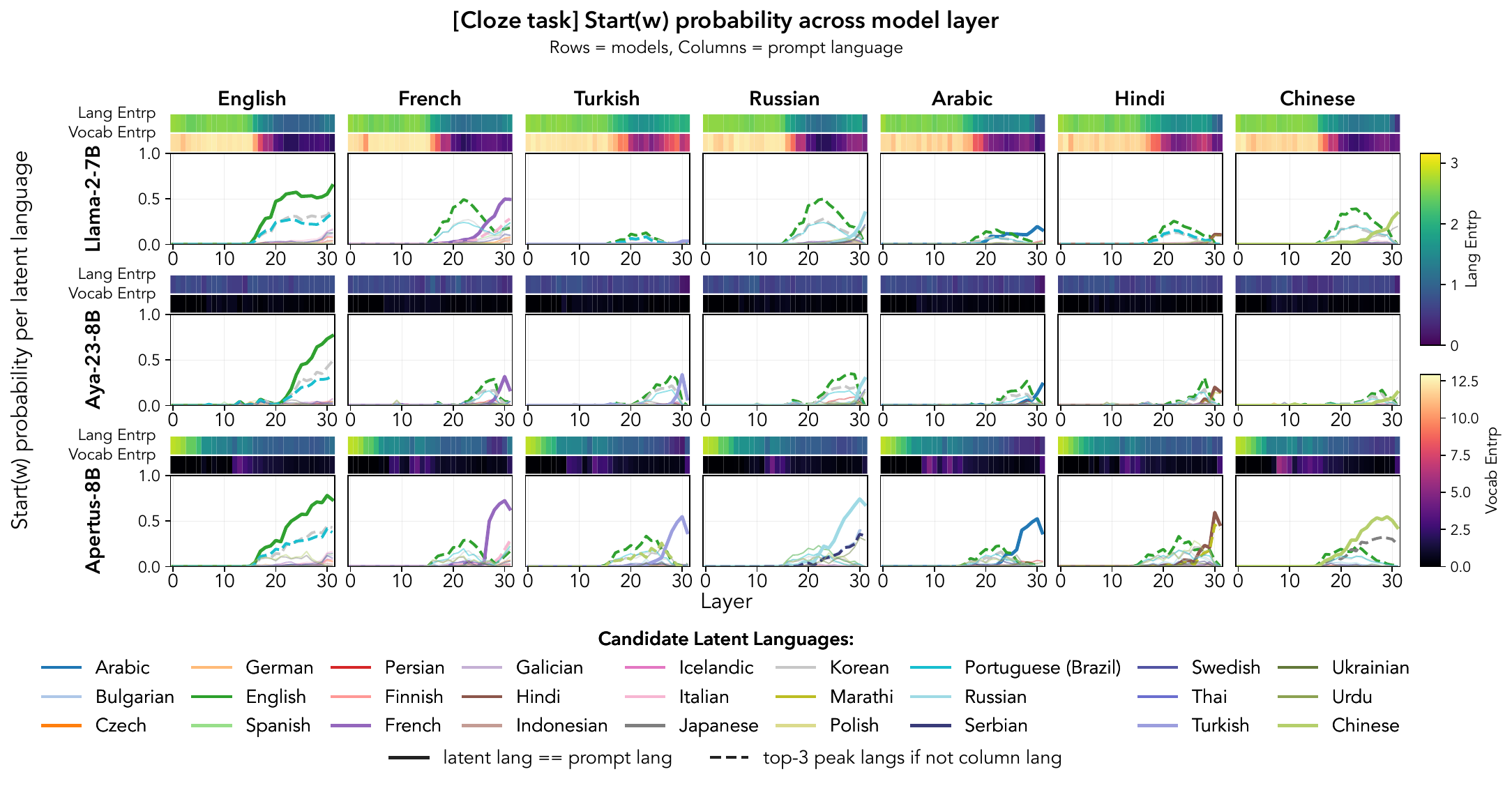}
    \caption{\textbf{Cloze prompts: latent language probability across model layers.} For each cloze prompt and model layer, we measure the summed next-token probability of all tokens that can begin the corresponding answer in each candidate language ($\startw{}$ probability). Rows are models and columns group prompts with the same prompt language. Upper strips show average latent language distribution entropy and average first-token vocabulary entropy.}
    \label{fig:app-cloze-startw}
\end{figure*}

\begin{figure*}[th]
    \centering
    \includegraphics[width=\textwidth]{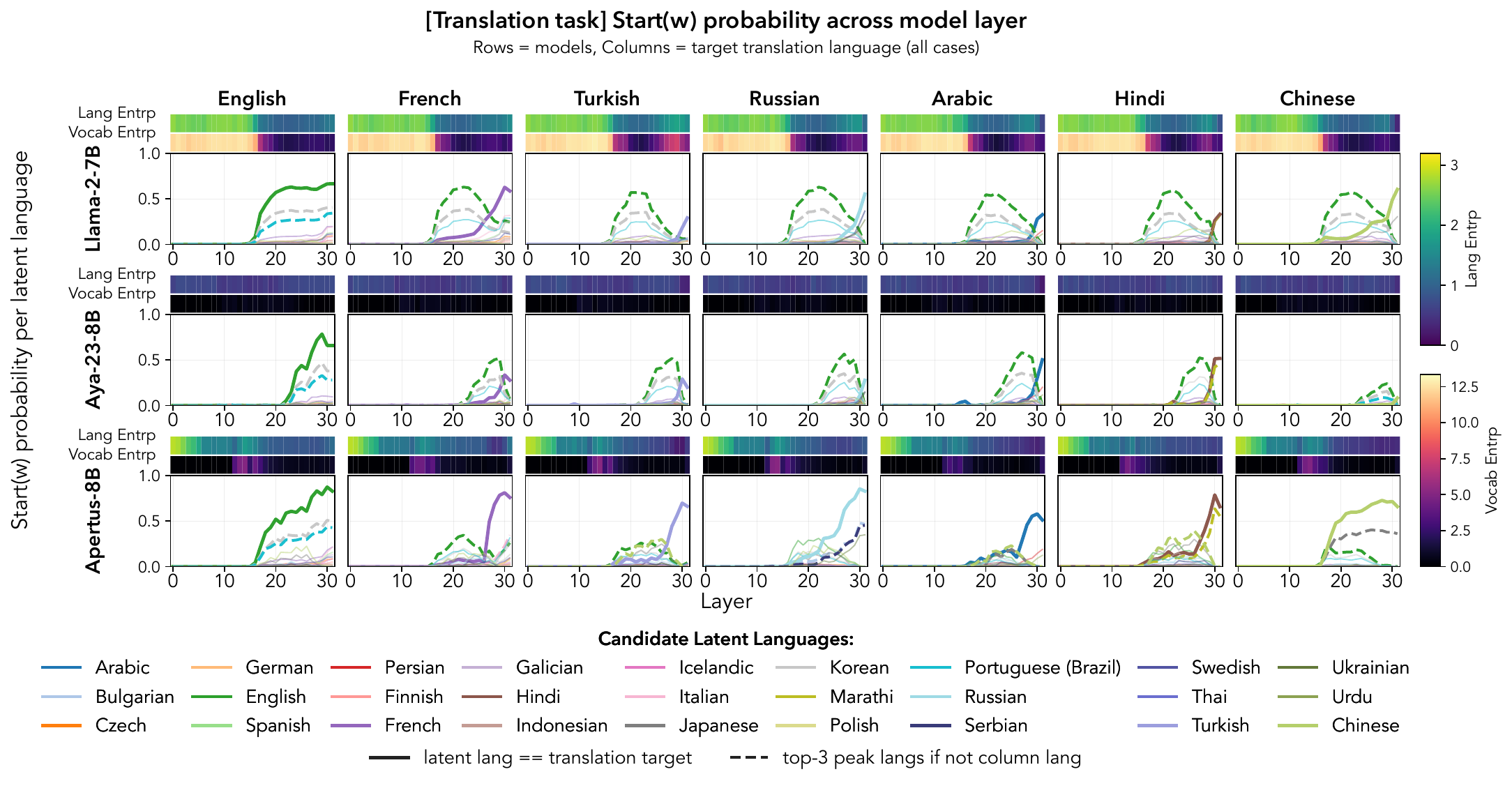}
    \caption{\textbf{Translation prompts: latent language probability across model layers, grouped by translation target.} For each translation prompt and model layer, we measure the summed next-token probability of all tokens that can begin the corresponding answer in each candidate language ($\startw{}$ probability). Rows are models and columns group prompts with the same translation target. English source and target cases are included. Upper strips show average latent language distribution entropy and average first-token vocabulary entropy.}
    \label{fig:app-translation-startw-by-target-all}
\end{figure*}

\begin{figure*}[th]
    \centering
    \includegraphics[width=\textwidth]{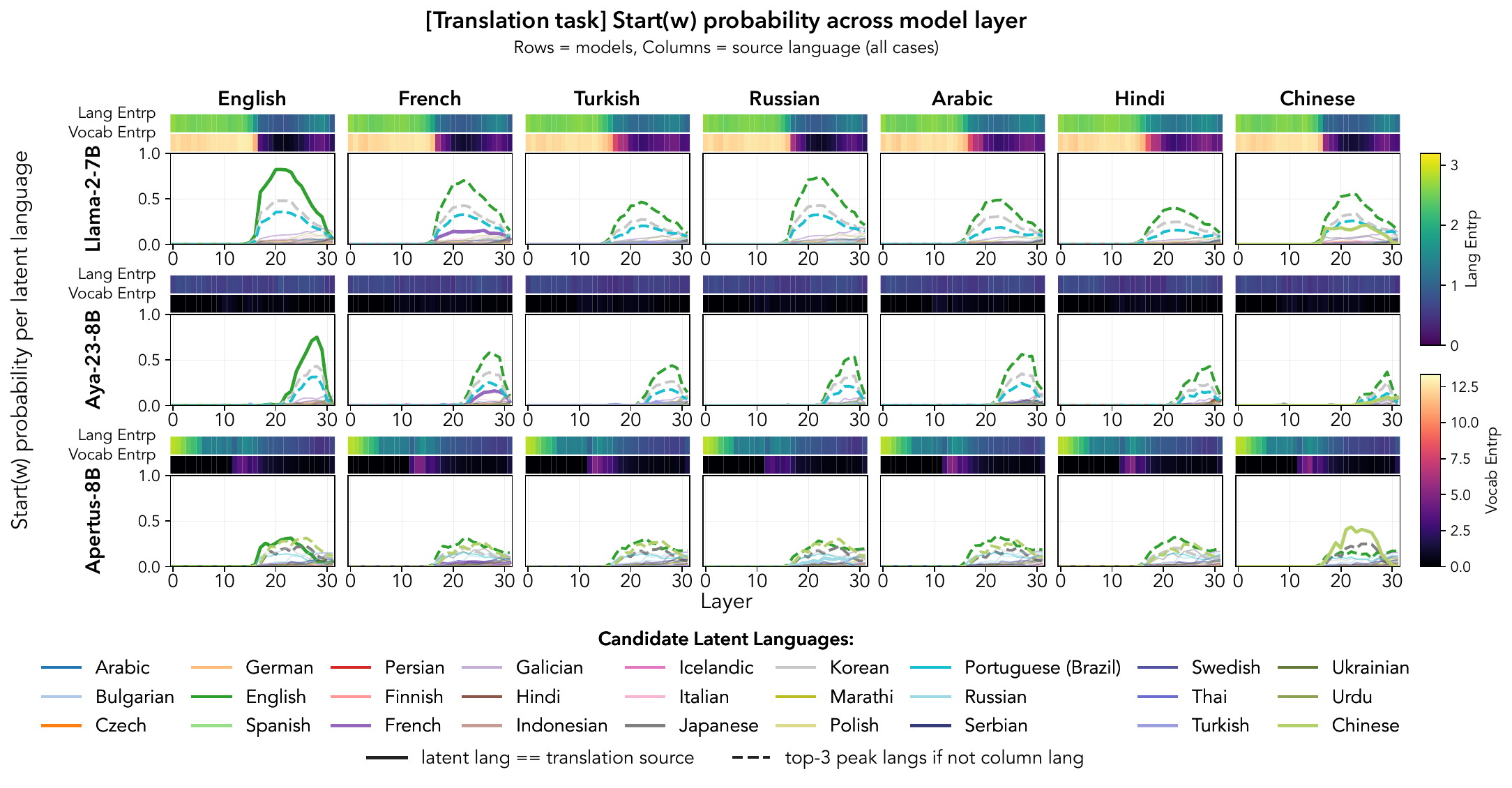}
    \caption{\textbf{Translation prompts: latent language probability across model layers, grouped by translation source.} For each translation prompt and model layer, we measure the summed next-token probability of all tokens that can begin the corresponding answer in each candidate language ($\startw{}$ probability). Rows are models and columns group prompts with the same translation source. English source and target cases are included. Upper strips show average latent language distribution entropy and average first-token vocabulary entropy.}
    \label{fig:app-translation-startw-by-source-all}
\end{figure*}

\begin{figure*}[th]
    \centering
    \includegraphics[width=\textwidth]{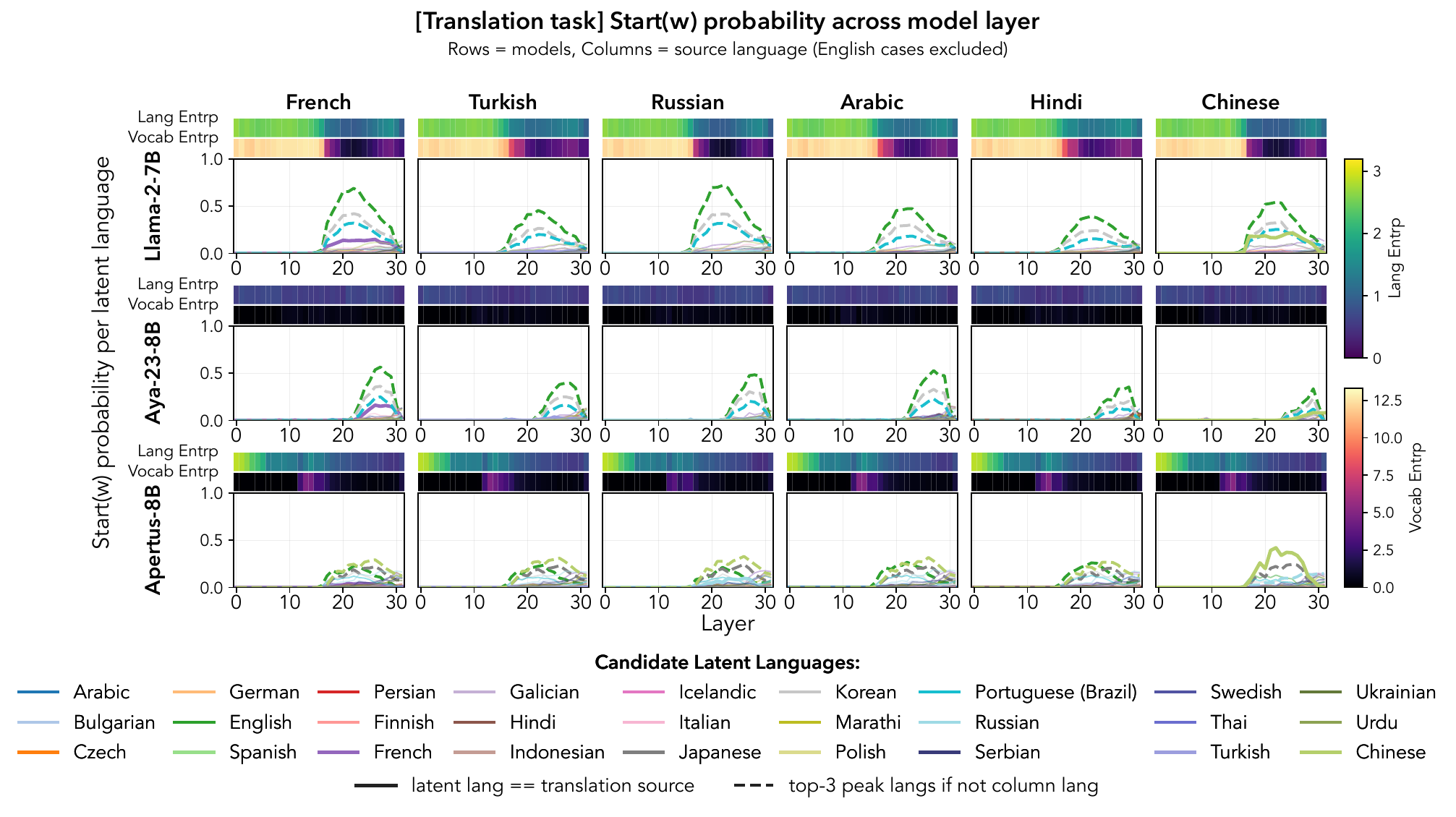}
    \caption{\textbf{Translation prompts: latent language probability across model layers, grouped by translation source.} For each translation prompt and model layer, we measure the summed next-token probability of all tokens that can begin the corresponding answer in each candidate language ($\startw{}$ probability). Rows are models and columns group prompts with the same translation source. English is excluded as a source or target, but retained as a candidate latent language. Upper strips show average latent language distribution entropy and average first-token vocabulary entropy.}
    \label{fig:app-translation-startw-by-source-no-en}
\end{figure*}

\begin{figure*}[th]
    \centering
    \includegraphics[width=\textwidth]{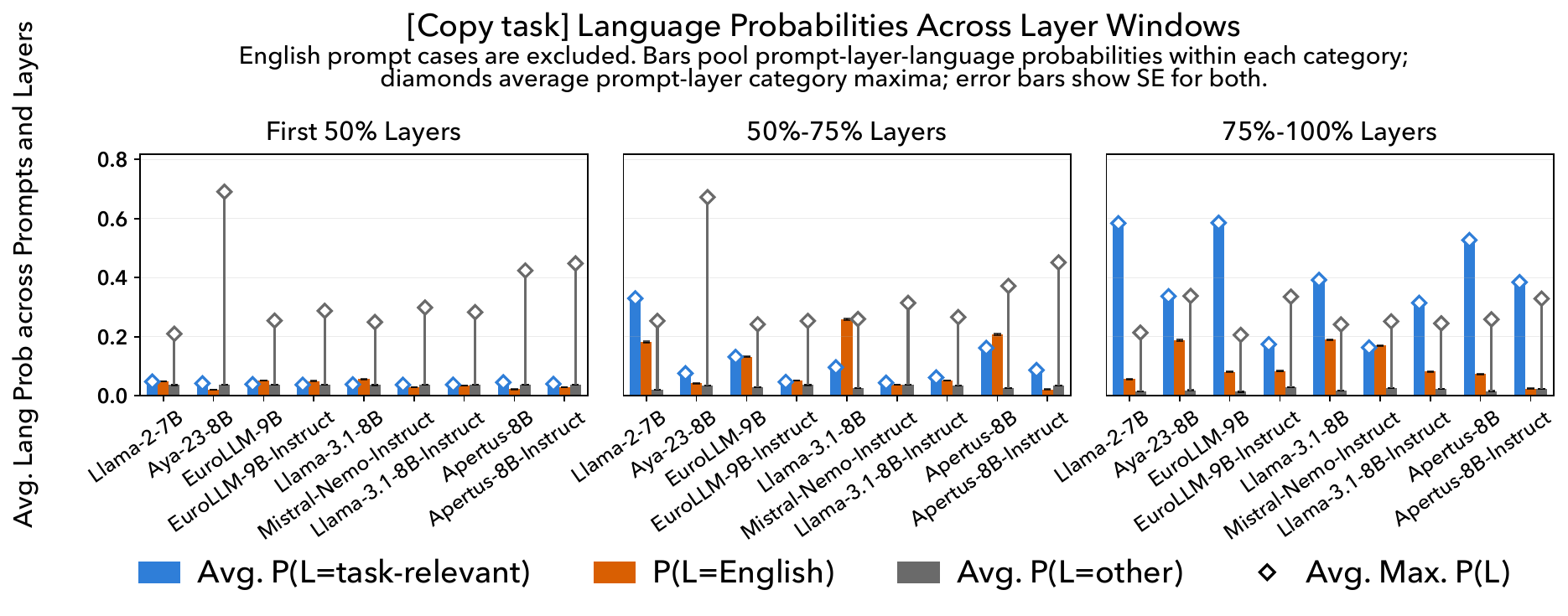}
    \caption{\textbf{Copy prompts: latent language probabilities across layer windows.} Columns show the first 50\%, 50--75\%, and 75--100\% of layers. Bars pool individual prompt--layer--language probabilities within each category before averaging; diamonds show the average maximum probability within the task-relevant or other-language category. English prompts are excluded, but English remains a candidate latent language. Error bars show SE.}
    \label{fig:app-copy-layer-windows}
\end{figure*}

\begin{figure*}[th]
    \centering
    \includegraphics[width=\textwidth]{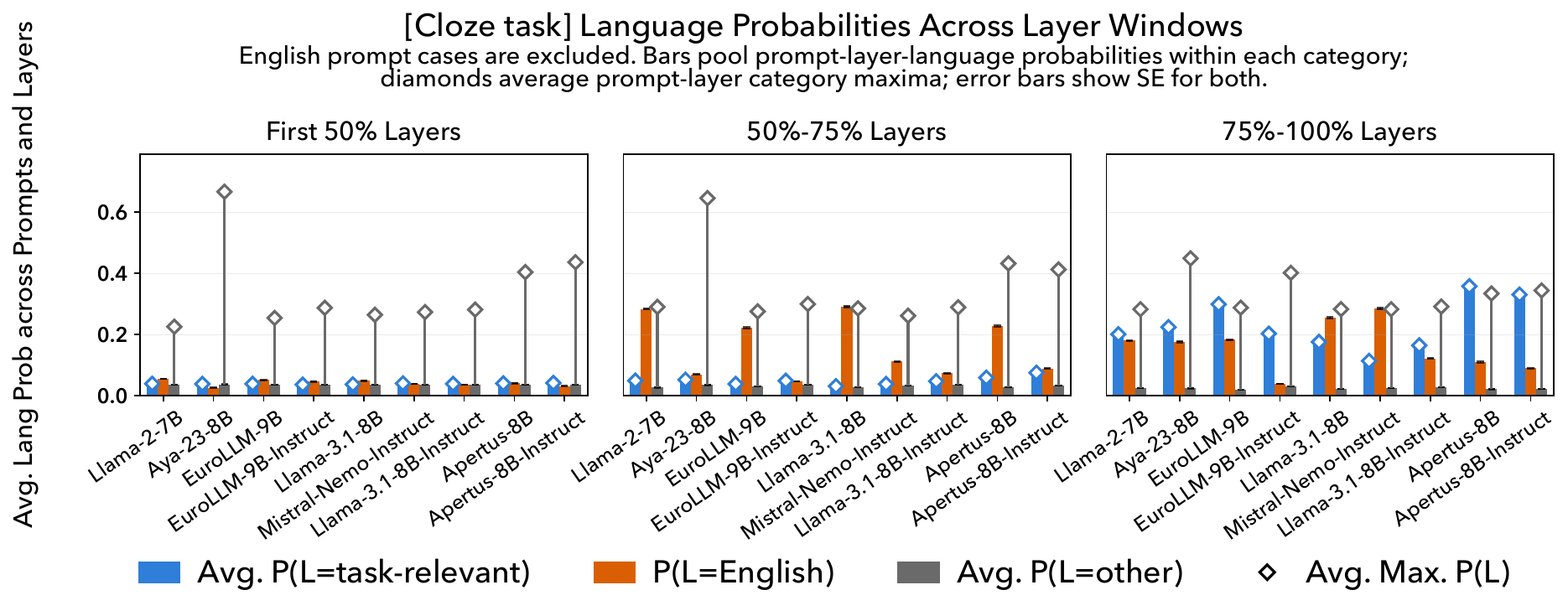}
    \caption{\textbf{Cloze prompts: latent language probabilities across layer windows.} Same layout and aggregation as Fig.~\ref{fig:app-copy-layer-windows}, evaluated on cloze prompts. English prompt cases are excluded.}
    \label{fig:app-cloze-layer-windows}
\end{figure*}

\begin{figure*}[th]
    \centering
    \includegraphics[width=\textwidth]{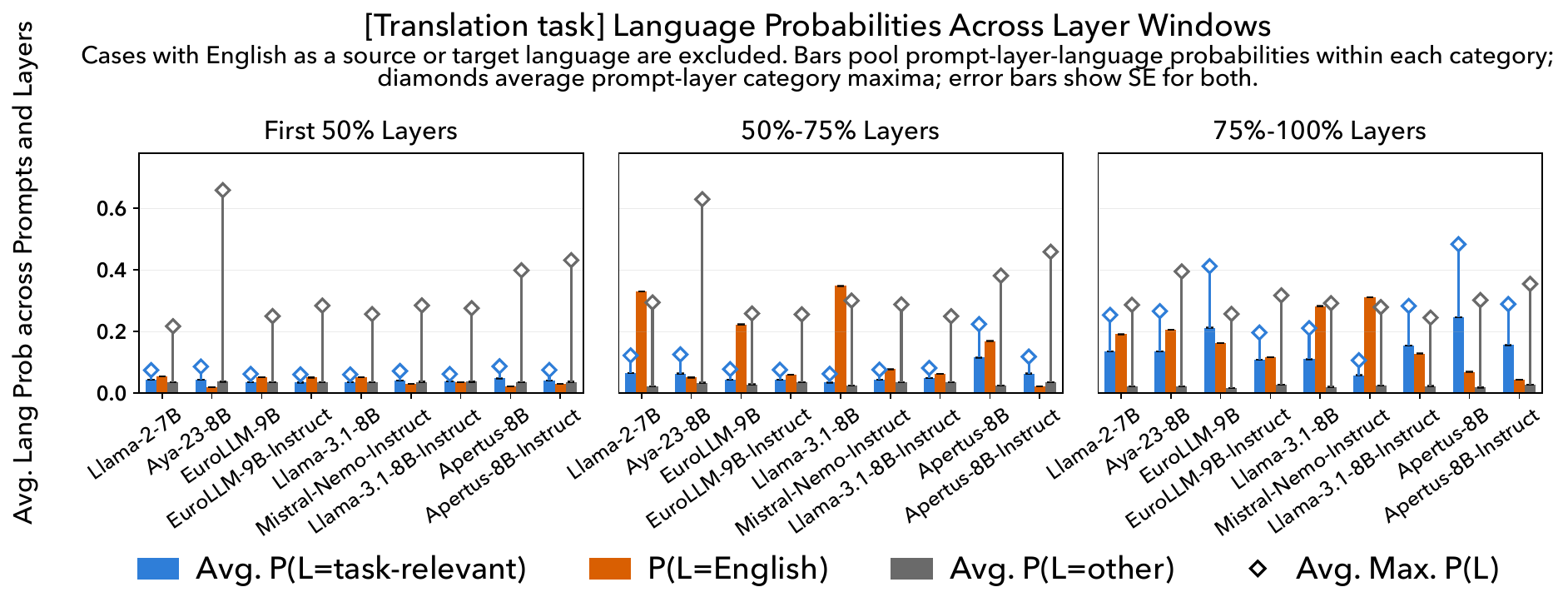}
    \caption{\textbf{Translation prompts: latent language probabilities across layer windows.} Same layout and aggregation as Fig.~\ref{fig:app-copy-layer-windows}. For translation, the task-relevant category contains both the translation source and target language. English is excluded as a source or target, but retained as a candidate latent language.}
    \label{fig:app-translation-layer-windows}
\end{figure*}

\begin{figure*}[th]
    \centering
    \includegraphics[width=\textwidth]{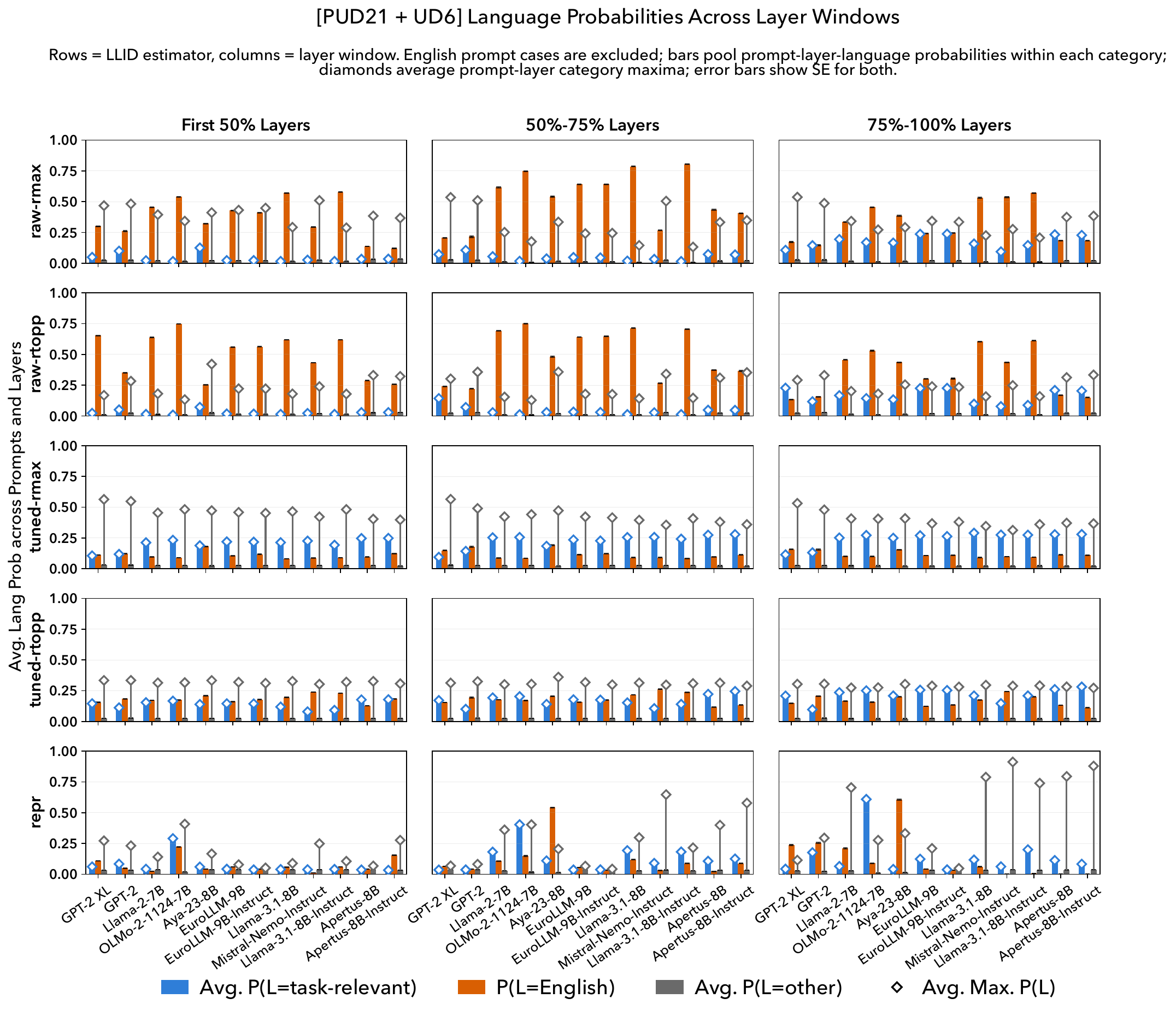}
    \caption{\textbf{Open-ended PUD21+UD6: latent language probabilities across layer windows.} Rows are LLID estimators and columns are layer windows. Bars, diamonds, and error bars use the same aggregation as Fig.~\ref{fig:app-copy-layer-windows}. The task-relevant category is the prompt language. English prompts are excluded, but English remains a candidate latent language.}
    \label{fig:app-pud-ud-layer-windows}
\end{figure*}

\begin{figure*}[th]
    \centering
    \includegraphics[width=\textwidth]{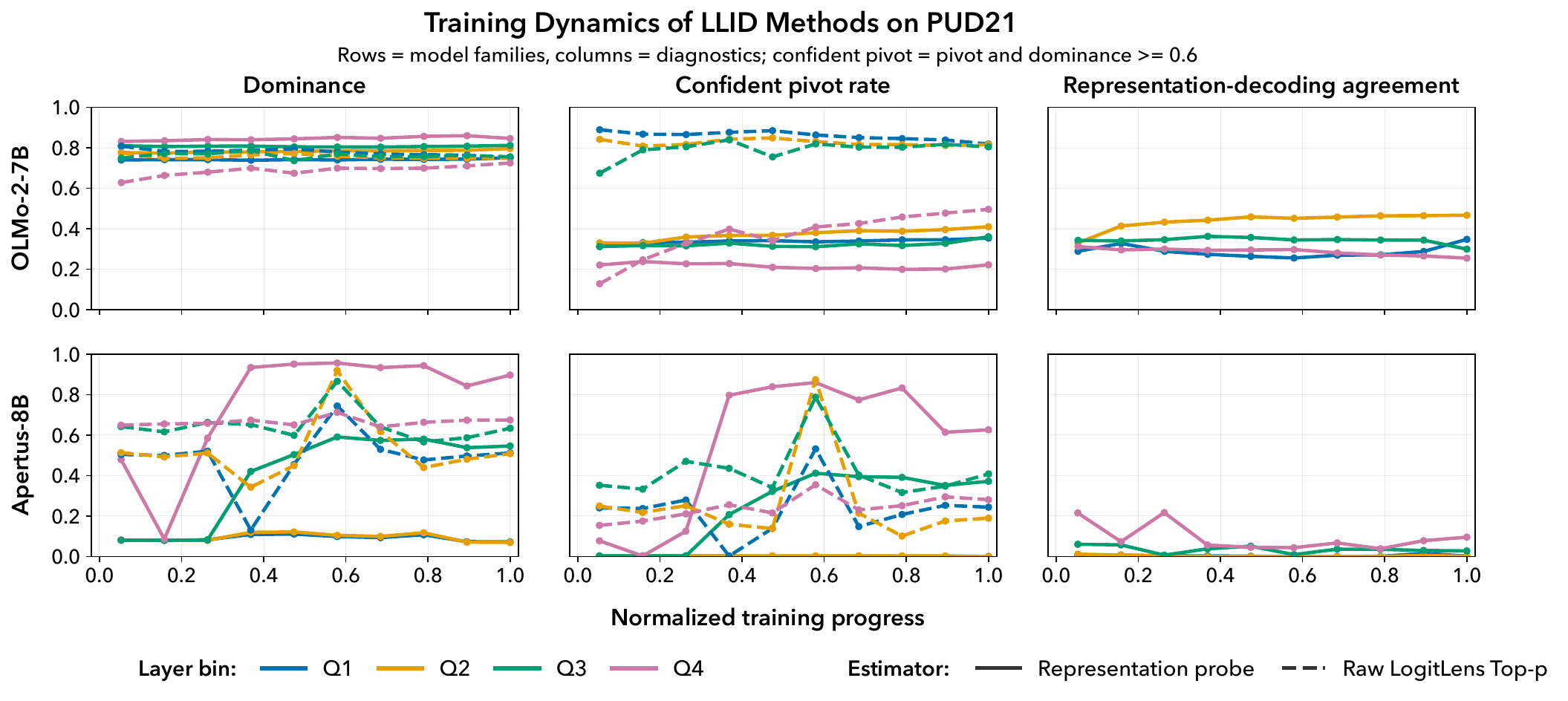}
    \caption{\textbf{Training dynamics of latent language estimates across pretraining on PUD21 with four layer bins.} This is the quartile-binned counterpart to Fig.~\ref{fig:training-dynamics}, using the same rows, metrics, estimator overlays, and prompt-level definitions. Colors divide the normalized layer index $j/J$ into quartiles: Q1 has $j/J < 0.25$, Q2 has $0.25 \leq j/J < 0.50$, Q3 has $0.50 \leq j/J < 0.75$, and Q4 has $j/J \geq 0.75$.}
    \label{fig:app-training-dynamics-quartiles}
\end{figure*}

\begin{figure*}[th]
    \centering
    \includegraphics[width=\textwidth]{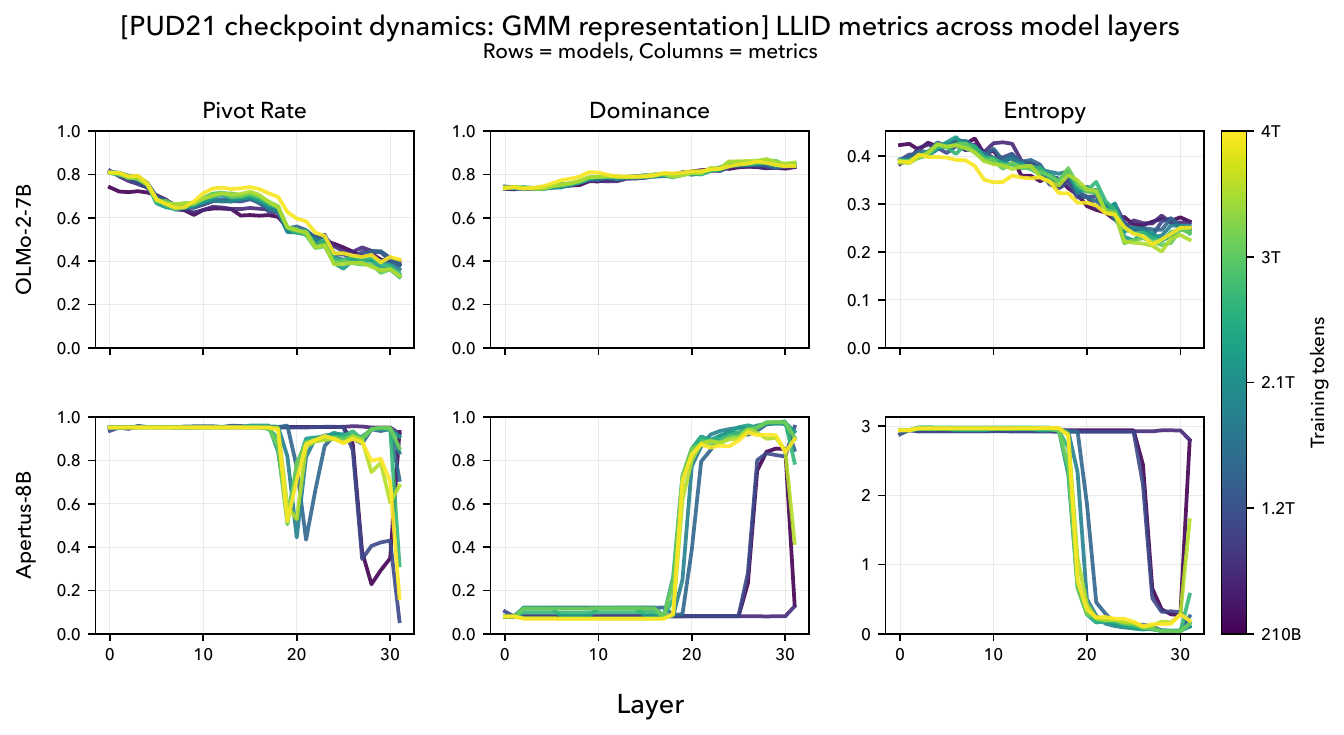}
    \caption{\textbf{Representation-based LLID checkpoint trajectories across layers on PUD21.} Rows are model families and columns are layerwise metrics. Each curve is one token-aligned checkpoint, colored by its number of training tokens. Metrics are averaged over prompts at each layer. Unlike Fig.~\ref{fig:training-dynamics}, this view keeps the full layer axis instead of aggregating layers into bins.}
    \label{fig:app-checkpoint-dynamics-repr}
\end{figure*}

\begin{figure*}[th]
    \centering
    \includegraphics[width=\textwidth]{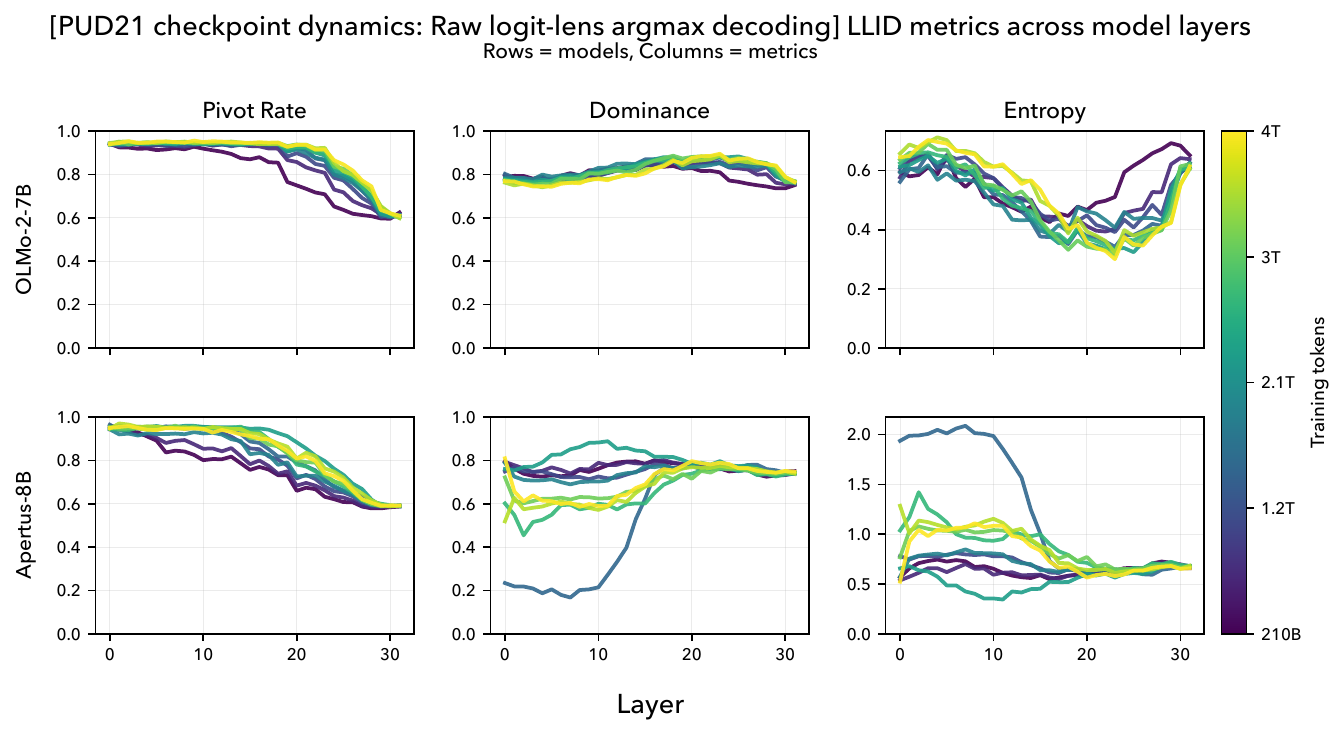}
    \caption{\textbf{Decoding-based LLID checkpoint trajectories across layers on PUD21.} Same layout as Fig.~\ref{fig:app-checkpoint-dynamics-repr}, using the raw logitlens argmax decoding probe instead of the representation probe.}
    \label{fig:app-checkpoint-dynamics-decoding}
\end{figure*}

\begin{figure*}[th]
    \centering
    \includegraphics[width=\textwidth]{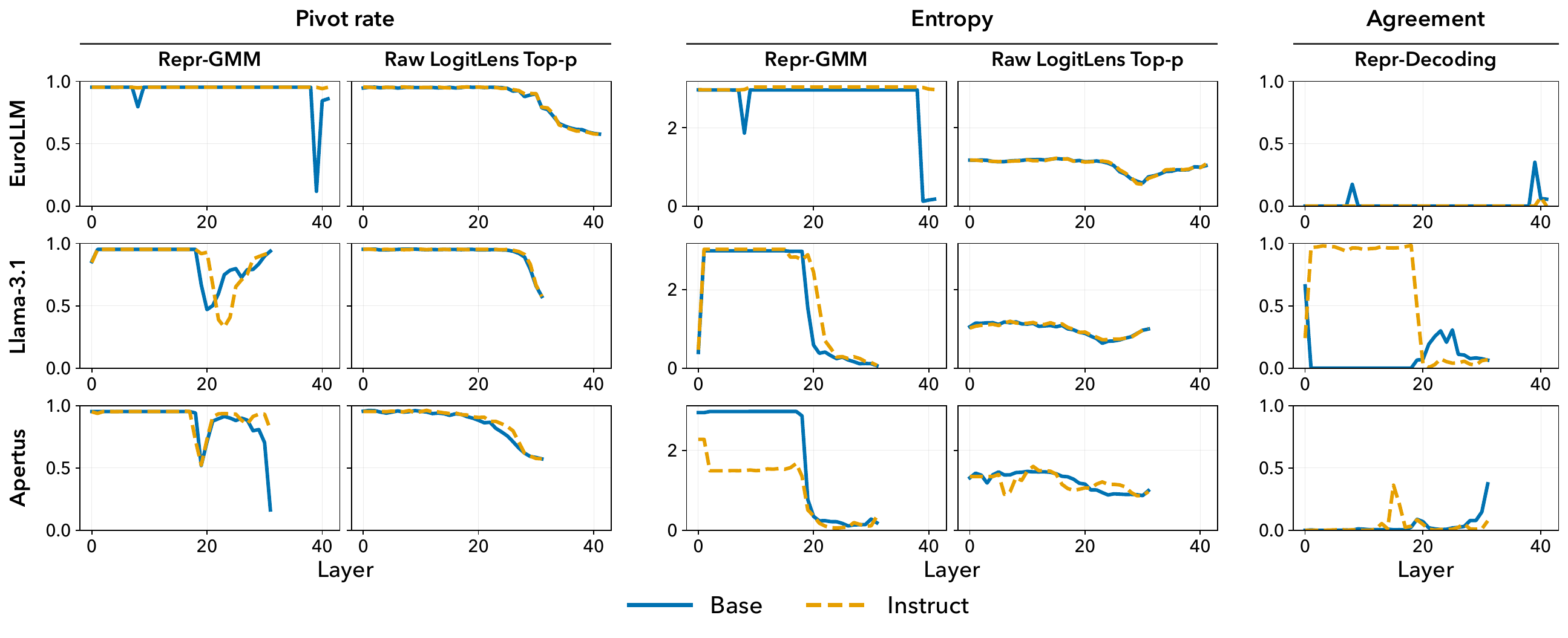}
    \caption{\textbf{Base-vs-instruct LLID trajectories on PUD21.} Rows are model families. Columns show pivot rate under \repr{} and raw logitlens \dectopp{}, entropy under the same two estimators, and their dominant language agreement. Blue solid and orange dashed curves denote base and instruction-tuned variants, respectively.
    Agreement is the fraction of matched prompt--layer pairs for which the two probes select the same dominant language.}
    \label{fig:app-base-vs-instruct-pud}
\end{figure*}

\begin{figure*}[th]
    \centering
    \includegraphics[width=\textwidth]{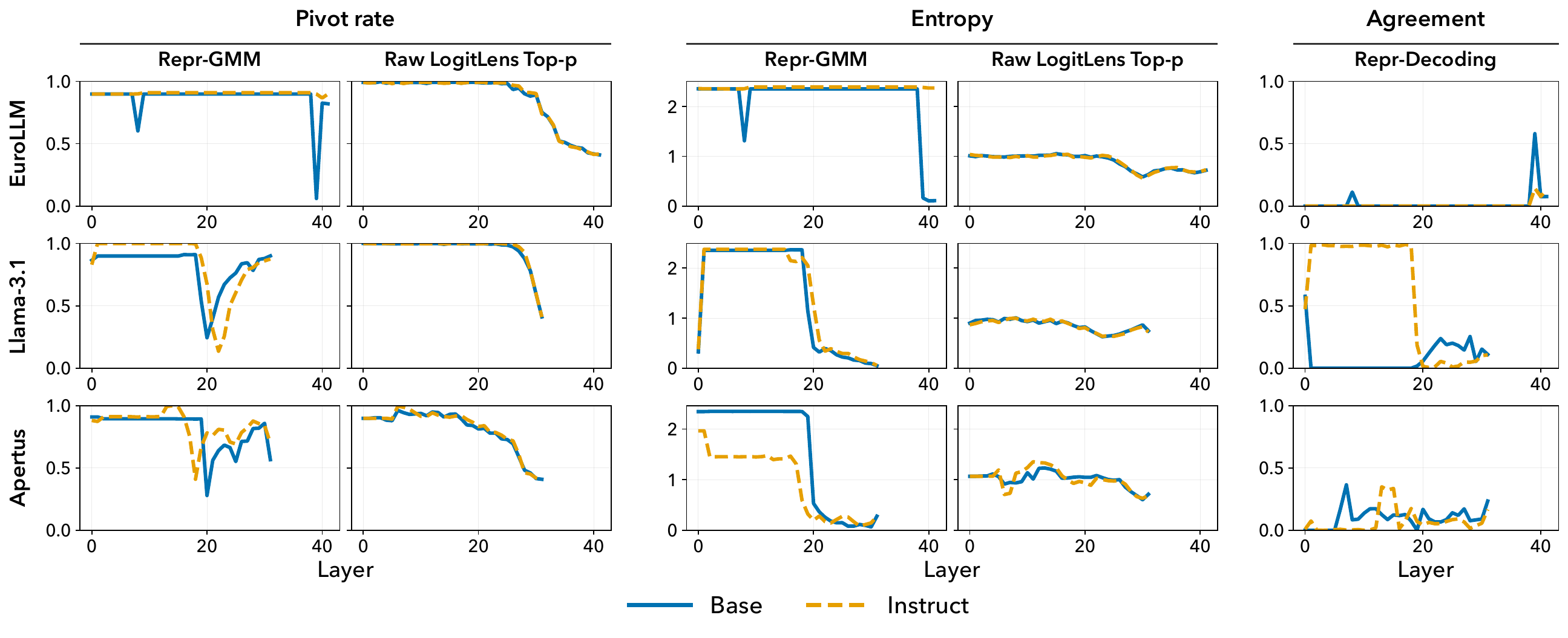}
    \caption{\textbf{Base-vs-instruct LLID trajectories on INCLUDE-10 with English as a candidate language.} Same layout as Fig.~\ref{fig:app-base-vs-instruct-pud}, evaluated on INCLUDE-10. INCLUDE contains no English prompts, but English is retained as a candidate latent language.}
    \label{fig:app-base-vs-instruct-include}
\end{figure*}

The appendix figures below are organized by analysis group. For the controlled synthetic tasks, Figs.~\ref{fig:app-copy-startw}--\ref{fig:app-translation-startw-by-source-no-en} show layerwise $\startw{}$ probability curves, while Figs.~\ref{fig:app-copy-layer-windows}--\ref{fig:app-translation-layer-windows} summarize the same task family into language-probability categories across layer windows. For open-ended language probabilities, Fig.~\ref{fig:app-pud-ud-layer-windows} gives the PUD21+UD6 counterpart. For training analyses, Figs.~\ref{fig:app-training-dynamics-quartiles}--\ref{fig:app-checkpoint-dynamics-decoding} show checkpoint dynamics and full-layer trajectories, while Figs.~\ref{fig:app-base-vs-instruct-pud}--\ref{fig:app-base-vs-instruct-include} compare base and instruction-tuned model variants.

\begin{figure*}[th]
    \centering
    \begin{subfigure}{0.49\linewidth}
        \centering
        \includegraphics[width=\textwidth]{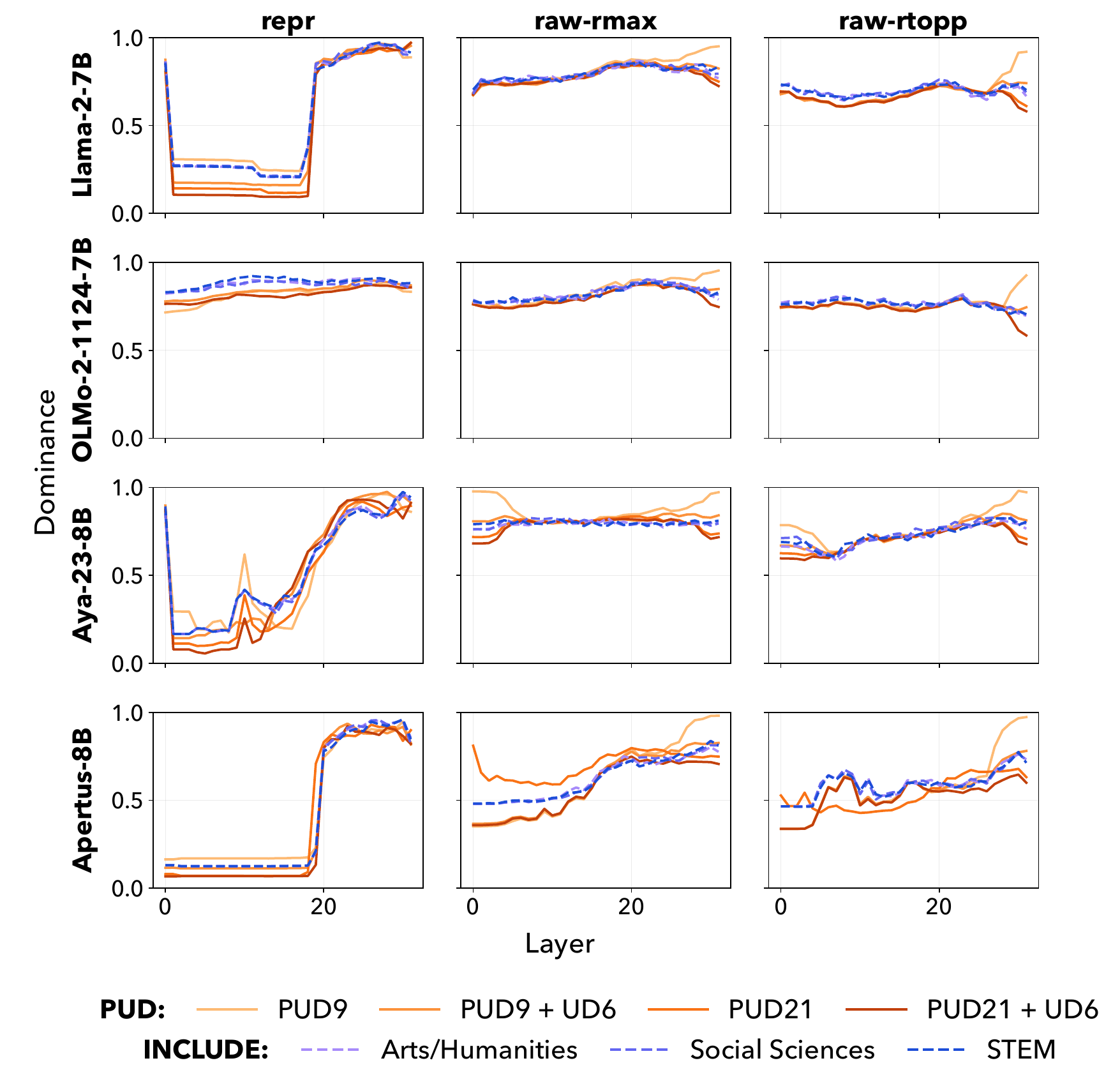}
        \caption{\textbf{Dominance}}
    \end{subfigure}
    \hfill
    \begin{subfigure}{0.49\linewidth}
        \centering
        \includegraphics[width=\textwidth]{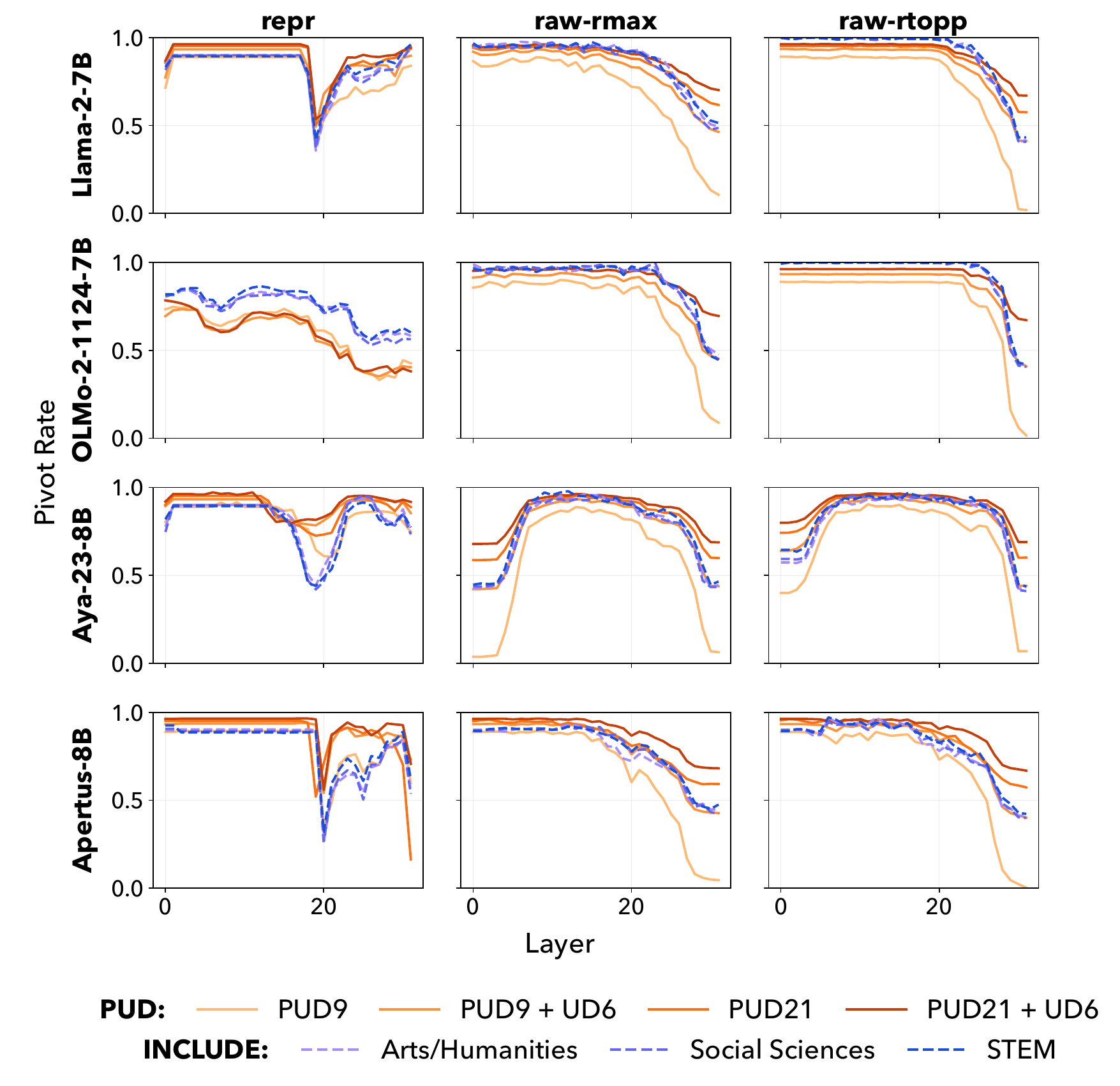}
        \caption{\textbf{Pivot rate}}
    \end{subfigure}
    \caption{\textbf{Latent language dominance and pivoting are comparatively stable across domains and language inventories.} Rows are language models and columns are LLID methods. For each prompt and layer, dominance is the highest probability assigned to a candidate latent language; a pivot occurs when the dominant language differs from the prompt language. Curves largely cluster across PUD variants (PUD9, PUD9+UD6, PUD21, PUD21+UD6) and INCLUDE domains (Arts/Humanities, Social Sciences, and STEM) within each model--estimator panel, while the layerwise trajectories can differ substantially across models.}
    \label{fig:app-domain-dominance-pivot}
\end{figure*}

\end{document}